\documentclass[conference,compsoc]{IEEEtran}
\usepackage{amssymb}
\usepackage{url}
\usepackage{stfloats}
\usepackage{hyperref}
\usepackage{placeins}
\usepackage{float}
\usepackage[T1]{fontenc} 
\usepackage{amsmath}
\usepackage{graphicx}
\usepackage{epstopdf}
\usepackage{amsfonts}
\usepackage{cite}
\usepackage{bm} 
\usepackage{mathrsfs}
\usepackage{pdfpages}
\usepackage{amsmath}
\usepackage{subfigure}
\usepackage{xcolor}

\begin{document}
\title{Adaptive Hybrid Optimizer based Framework for Lumpy Skin Disease Identification}

        \author{Ubaidullah$^1$, Muhammad Abid Hussain$^1$,
        	Mohsin Raza Jafri$^1$, Rozi Khan$^1$, \\ Moid Sandhu$^{2,3}$, Abd Ullah Khan$^{1,4}$, Hyundong Shin$^{4,*}$ \\
        \small{}
        $^1$Department of Computer Science, NUST Balochistan Campus, NUST, Pakistan (e-mail: obaidullah@nbc.nust.edu.pk, \\  abidmaharvi@hotmail.com,   mjafari@nbc.nust.edu.pk, rozi.phdcs24.nbc@student.nust.edu.pk)\\
        $^2$Australian e-Health Research Centre (AEHRC), CSIRO, Brisbane, Australia (e-mail: moid.sandhu@csiro.au)\\
        $^3$Queensland University of Technology, Brisbane, Australia\\
        $^4$CQI Lab, Department of Electronics and Information Convergence Engineering, Kyung Hee University, \\ Global Campus, South Korea (e-mail: \{abdullah, hshin\}@khu.ac.kr)\\
        *Corresponding author: Hyundong Shin}

\markboth{}%
{\MakeLowercase{}}

\maketitle

\begin{abstract}
Lumpy Skin Disease (LSD) is a contagious viral infection that significantly deteriorates livestock health, thereby posing a serious threat to the global economy and food security. Owing to its rapid spread characteristics, early and precise identification is crucial to prevent outbreaks and ensure timely intervention.  In this paper, we propose a hybrid deep learning-based approach called LUMPNet for the early detection of LSD. LUMPNet utilizes image data to detect and classify skin nodules -- the primary indicator of LSD. To this end, LUMPNet uses YOLOv11, EfficientNet-based CNN classifier with compound scaling, and a novel adaptive hybrid optimizer. More precisely, LUMPNet detects and localizes LSD skin nodules and lesions on cattle images. It exploits EfficientNet to classify the localized cattle images into LSD-affected or healthy categories. To stabilize and accelerate the training of YOLOv11 and EfficientNet hybrid model, a novel adaptive hybrid optimizer is proposed and utilized. We evaluate LUMPNet at various stages of LSD using a publicly available dataset. Results indicate that the proposed scheme achieves 99\% LSD detection training accuracy, and outperforms existing schemes. The model also achieves validation accuracy of 98\%. Moreover, for further evaluation, we conduct a case study using an optimized EfficientNet-B0 model trained with the AdamW optimizer, and compare its performance with LUMPNet. The results show that LUMPNet achieves superior performance.
\end{abstract} 

\IEEEpeerreviewmaketitle

\section{Introduction}
Precision livestock farming utilizes automated structures to monitor livestock, enhancing their production, reproduction, health, welfare, and environmental impact \cite{krampe2024designing}. It plays a vital role in the early detection of diseases and contamination in animals. Specifically, pores and skin parasitic infections pose a chief financial undertaking within the livestock enterprise. These infections cause discomfort, impairing the animal’s capacity to relax and eat well. Moreover, those infections also compromise the skin’s innate defense mechanisms, which in turn increase susceptibility to bacterial infections. Certain ectoparasites, which include blood-feeding flies and precise tick species, are economically significant as they transmit diverse illnesses. The pores and skin lesions attributable to those parasites are specifically harmful. 

\begin{table*}
\centering
\caption{Diseases in Cattle with Skin Nodules and Detection Metrics}
\begin{tabular} {|p{1.7cm}|p{3cm}|p{1.8cm}|p{2.5cm}|p{3cm}|p{2.5cm}|}
\hline
\textbf{Disease} & \textbf{Skin Nodules Description} & \textbf{Fatality Rate} & \textbf{Early Detection Indicators} & \textbf{Other Metrics} & \textbf{Nodules Detection Importance} \\ \hline
Lumpy Skin Disease (LSD) & Severe nodules on the skin, mainly on the neck and back & High & Fever, swelling, reduced milk yield & Requires early diagnosis to prevent spread & Very high \\ \hline
Bovine Tuberculosis (TB) & Granulomas (nodules) may appear on skin or internal organs & Low & Coughing, weight loss, and fever & Skin nodules often internal, may require advanced tests & Moderate \\ \hline
Poxvirus Infections & Nodular lesions, usually around the head or genital area & Low to Moderate & Skin lesions, scabs, fever & Can affect other animals; highly contagious & Moderate \\ \hline
Bluetongue & Nodules around the head, ears, and neck & Moderate & Swelling, drooling, fever, lameness & Often seasonal, transmitted by vectors & High \\ \hline
Contagious Ecthyma (Orf) & Scab-like lesions around mouth, udder, and teats & Low & Lesions around the mouth or udder & Highly infectious, zoonotic & High \\ \hline
Mycotic Dermatitis & Nodules and crusts formed due to fungal infection & Low & Scaly skin, itching, hair loss & Can spread through contaminated environments & Moderate \\ \hline
Schmallenberg Virus (SBV) & Skin nodules with fever and swelling of joints & Low (usually) & Fever, swelling, reduced appetite & Vector-borne, affects pregnant cattle & High \\ \hline
Bovine Parapoxvirus & Nodules on the udder, teats, and skin & Low & Lesions on udder and teats & Zoonotic, easily spread in unsanitary conditions & High \\ \hline
\end{tabular}
\label{tab_diseases}
\end{table*}

\begin{figure}[t]
\centering
\includegraphics[height = 6cm, width = 7.5cm]{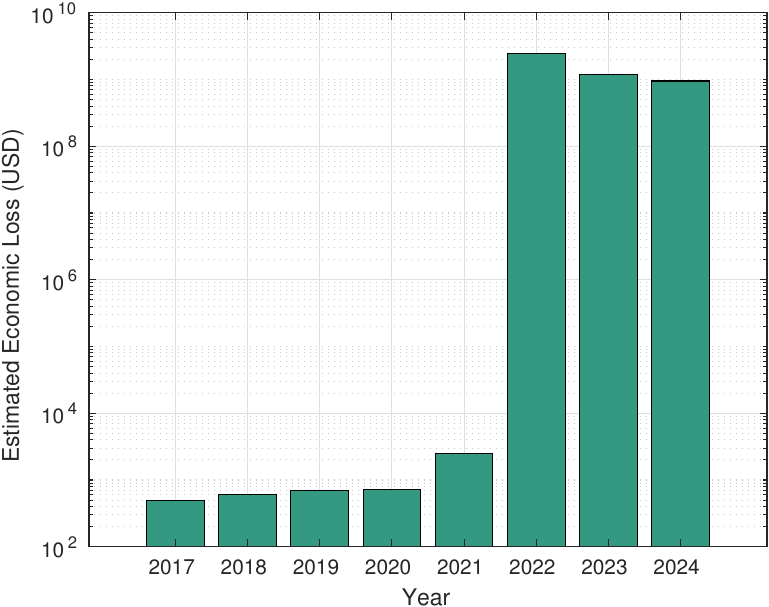}
\caption{Estimated Economic Loss due to Lumpy Skin Disease (2017–2024) \cite{naidu2025assessment} \label{tab:losses}}
\end{figure}

Livestock, consisting of farm animals, cows, and buffalo, are vulnerable to a viral infection referred to as Lumpy Skin Disease (LSD) ~\cite{tuppurainen2018introduction}. This disease is transmitted by means of blood-feeding insects like ticks, flies, and mosquitoes. In addition to causing fever and skin nodules, the virus can be fatal, especially for animals that have never been exposed to it before. Control measures include vaccination and culling infected animals. LSD results in considerable financial losses, affecting species such as cows and water buffalo. The disease is caused by the LSD virus, which belongs to the Capripox genus within the Poxiviridae family. The virus persists in lesions or scabs for prolonged intervals, making these skin lesions the primary source of transmission. Cattle farms around the world experience high infection rates, with over 7,500 deaths connected to LSD in the past 5 years, out of 190,000 reported cases as in Figure. \ref{tab:losses}. LSD critically affects the economic system, leading to reduced milk production and animal losses. The Table ~\ref{tab_diseases} summarizes the diverse diseases and their signs related to livestock. 
There are several cattle diseases associated with skin nodules, which can be classified based on key characteristics such as the description of nodules, fatality rate, early detection indicators, and the importance of nodule detection. LSD stands out because of its high fatality rate and the extreme nature of the skin nodules, which can be commonly observed on the neck and lower back. Early detection of LSD is crucial to save from its rapid unfold, with signs such as fever, swelling, and reduced milk yield serving as key indicators. While different sicknesses, which include Bovine Tuberculosis, Bluetongue, and Contagious Ecthyma, additionally contain skin nodules, LSD requires mainly urgent interest because of its significant impact on farm animals' health and productivity. Adamu et al. \cite{adamu2024lumpy} conducted an experimental analysis of LSD‑affected cattle in Ethiopia. They confirmed LSDV as the primary cause of the outbreak and identified nodules and fever as the main symptoms of the disease.

Recent advancements in artificial intelligence (AI) have also made it possible to perform the early detection of various livestock diseases. In this direction, various researchers have reported their work in the literature related to preciosn livestock farming. Chong et al. \cite{he2023enhanced} propose a method for estimating sheep live weight using LiteHRNet, a Lightweight High-Resolution Network, with RGB-D images. The design of efficient network heads was guided by Class Activation Mapping (CAM), ensuring visual interpretability and applicability in real-world environments. The technique was tested on a challenging dataset together with 726 RGB-D pictures of sheep, with weights starting from 19.5 to 94 kg. The consequences of the comparative experiments show that the lightweight Convolutional Neural Network (CNN) model trained on RGB-D pictures yields an acceptable weight estimation overall performance. It successfully achieves a Mean Average Percentage Error (MAPE) of 14.605\%, with only 1.06 M parameters. Gouda eta l. \cite{gouda2025comparative} predicted LSD using multiple ML models under multiclass balance. They apply under-sampling and oversampling to address the class imbalance. Random forest along with random oversampling achieved 82\% accuracy.

Qiao et al. \cite{qiao2022c3d} introduce a deep learning framework designed to monitor and classify dairy cow behaviors. This framework combines the Convolutional 3D (C3D) network with Convolutional Long Short-Term Memory (ConvLSTM). The system classifies five key behaviors: feeding, exploring, grooming, walking, and standing. The model initially extracts 3D convolutional features from the video frames using the C3D network. In the second phase, the scheme captures spatio-temporal information using ConvLSTM. This information is then passed through a softmax layer for classification. Using a 30-frame video length, the proposed method achieved classification accuracies of 90.32\% and 86.67\% on datasets for calves and cows, respectively, outperforming state-of-the-art models such as BiLSTM, Inception-V3, SimpleRNN, LSTM, and C3D.


\begin{figure*}
\centering
\includegraphics[width=19cm, height=14cm, trim=8pt 2pt 2pt 2pt, clip]{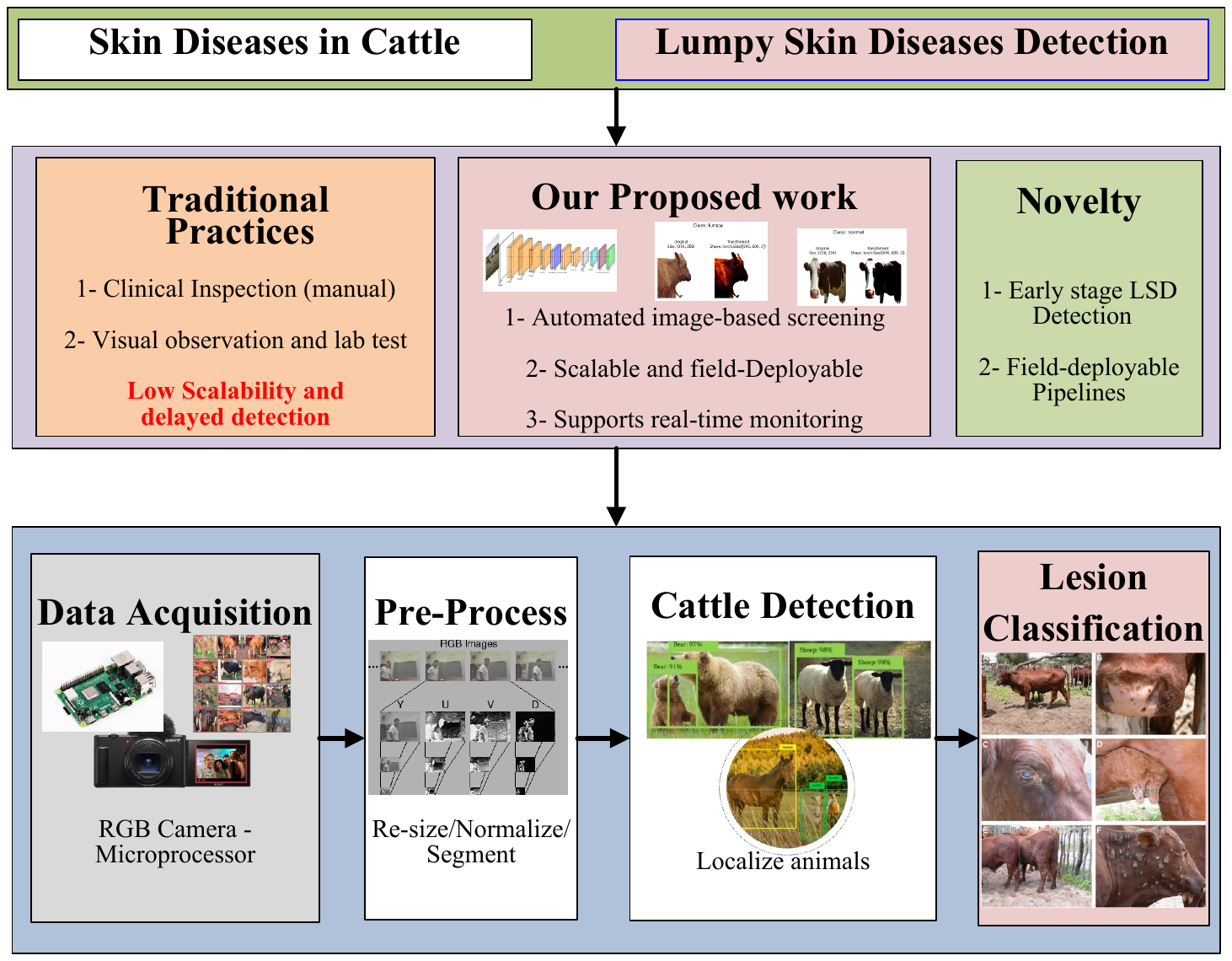}
\caption{Summary of the Proposed work: LUMPNet \label{abstract}}
\end{figure*}

In another study, Qiao et al. \cite{qiao2021intelligent} provide an in-depth analysis of key techniques in precision livestock farming, focusing on methods for cattle identification, live weight estimation, and body condition evaluation. They review multiple applicable studies and present them in a comprehensive and organized manner. Based on the findings and evaluation of latest improvements, the authors are expecting that the adoption of high-precision, non-contact, and automatic technologies may be critical for the destiny of precision cattle farming. They additionally emphasize the significance of integrating emerging 3D model reconstruction and deep learning methods into these fields of study.

Wang et al. \cite{wang2023computer} develop a system for monitoring the respiratory rate in group-housed pigs. The proposed system identifies video clips where the pigs are in a resting state. It applies an object detection algorithm to automatically detect each animal and select the region of interest (ROI) without requiring manual input. They estimate the respiratory rate by analyzing time-varying features from the ROI. They collect video footage of a group of five pigs wearing an ECG belt around their abdomens to provide gold-standard respiratory rate measurements. The comparison of the estimates from the AI-based method showed a high level of accuracy, with a mean absolute error of 2.38 breaths per minute. Saqib et al. \cite{muhammad2024lumpy} employ the MobileNetV2 model and the RMSprop optimizer in order to devise a novel deep learning scheme for detecting LSD. They test their scheme on both healthy and infected cattle datasets, achieving 95\% accuracy. They outperform existing benchmarks by 5–10\%. The image dataset includes 464 images of healthy cows and 329 images of LSD-affected cows. 

Xiao et al. \cite{xiao2022cow} propose a method for identifying individual cows using top-view images. They capture a top-down image of the cow, followed by the application of an enhanced Mask R-CNN model to segment the image and extract features related to the cow's back shape. They employ the Fisher method to select the optimal subset of features. They use a support vector machine (SVM) classifier for cow identification. Compared to the traditional Mask R-CNN model, the improved Mask R-CNN model achieves precision, recall, F1 score, and average precision values of 98.21\%, 96.48\%, 97.34\%, and 97.39\%, respectively. Mahmud et al. \cite{mahmud2021systematic} summarize the details of recent advancements in the application of deep learning in precision cattle farming. They specifically emphasize health monitoring and identification. Out of 56 selected studies, 58\% focused on cattle identification, while the remaining studies concentrated on health monitoring. The analysis revealed the use of 20 different deep learning models, with CNNs being the most commonly utilized, followed by LSTM, Mask-RCNN, and Faster-RCNN models.

\begin{table}
\caption{Stages and Symptoms of LSD}
\begin{tabular} {|p{2.5cm}|p{5.1cm}|}
\hline
\textbf{Stage} & \textbf{Symptoms} \\
\hline
\textbf{Incubation Stage} & - No visible symptoms, virus replicates. \\
 & - Usually lasts 2-4 weeks after infection. \\
\hline
\textbf{Early Stage} & - Fever. \\
 & - Loss of appetite. \\
 & - Depression. \\
 & - Mild respiratory distress. \\
 & - Lesions start to appear. \\
\hline
\textbf{Intermediate Stage} & - Development of large lesions on the skin. \\
 & - Lesions range from 2-5 cm in size. \\
 & - Lesions start on the neck, head, and limbs. \\
\hline
\textbf{Progression Stage} & - Lesions ulcerate and secrete fluid. \\
 & - Increased body temperature continues. \\
 & - Secondary bacterial infections may occur. \\
 & - Lameness due to lesions. \\
\hline
\end{tabular}
\label{lsd_stages}
\end{table}

\begin{table*}
    \centering
    \caption{Comparison of Our Work with Related Works in LSD Detection and Monitoring}
    \begin{tabular}{|p{1.5cm}|p{2.7cm}|p{2.7cm}|p{2.7cm}|p{2.7cm}|p{2.7cm}|}
        \hline
        \textbf{Ref. (Year)} & \textbf{Approach} & \textbf{Model Performance} & \textbf{Dataset} & \textbf{Evaluation Methods} & \textbf{Challenges} \\
        \hline
        Girma et al. \cite{girma2021identify} (2021) & CNN for feature extraction and SVM for classification in LSD detection. & 95.7\% accuracy with SVM for LSD classification. & 1,740 image dataset for LSD detection. & Evaluated with SVM, Random Forest with SVM achieving 95.7\% accuracy. & Noise removal techniques needed for better image quality and region identification. \\
        \hline
        Neha et al. \cite{ujjwal2022exploiting} (2022) & Random Forest for LSD prediction in specific geographic locations. & 97\% accuracy with Random Forest for LSD prediction. & 18,603 instances and 16 features from a dataset. & Performance comparison with Random Forest (97\% accuracy). & Image noise and the challenges of accurate segmentation for LSD detection. \\
        \hline
        Kumar et al. \cite{senthilkumar2024early} (2024) & Automated LSD detection using deep learning with transfer learning & VGG16 and MobileNetV2 achieved accuracies of 96.07\% and 96.39\% & Publicly available datasets of healthy and LSD-affected cattle & Performance evaluation using accuracy, sensitivity, specificity, precision & Challenges in dataset balancing and the need for high sensitivity. \\
        \hline
        Saqib et al. \cite{muhammad2024lumpy} (2024) & Deep learning approach using MobileNetV2 and RMSprop optimizer. & 95\% accuracy with MobileNetV2 for LSD detection. & 464 healthy and 329 LSD-affected cattle images. & 95\% accuracy on healthy and LSD-affected cattle images dataset. & Partitioning dataset to balance the classes and ensuring diversity in images. \\
        \hline
        Asad etal. \cite{ullah2025vision} (2025) & ViT classifier with preprocessing (resizing, normalization, augmentation) & 98\% accuracy  & 8,000 cattle images & Evaluated using accuracy, precision, recall, and F1 score. & Ensuring generalization to field conditions \\
        \hline
        \textbf{Proposed: LUMPNet} (2025) & \textbf{YOLOv11 + EfficientNet + AWDR for LSD early detection and monitoring.} & \textbf{99\% accuracy in early detection of LSD.} & \textbf{1024 image dataset, 324 for LSD detection} & \textbf{Multiple-stage performance assessment with 99\% accuracy.} & \textbf{Handling early-stage lesion variability, ensuring precise region localization} \\
        \hline
    \end{tabular}
        \label{tab:comparison1}
\end{table*}
LSD is characterized by fever, swollen superficial lymph nodes, and numerous nodules ranging from 2 to 5 centimeters in diameter on the skin and mucosal membranes ~\cite{al2014lumpy}. Infected cattle also suffer from lameness and swelling in their limbs. The virus often causes lasting skin damage in affected animals. The disease leads to poor growth, infertility, abortion, chronic weakness, decreased milk production, and even death. Fever typically appears about a week after the viral infection, with the initial fever possibly exceeding 41°C and lasting for up to a week. During this period, all superficial lymph nodes become enlarged. The distinctive nodules appear 7 to 19 days after infection, accompanied by mucopurulent discharge from the nose and eyes. These nodules affect deeper layers of skin, such as the subcutis and even the muscle, impacting both the dermis and epidermis. The lesions may be well-defined or merge together and can appear anywhere on the body, though they are most commonly found on the head, neck, udder, scrotum, vulva, and perineum. Skin lesions may either resolve quickly or persist as painful lumps. In some cases, the lesions may become more severe, forming deep ulcers that may be pus-filled and surrounded by granulation tissue. Initially, the nodules discharge serum and appear gray to white upon cutting, eventually developing a central necrotic core after approximately two weeks. The virus is more easily transmitted through the nodules on mucous membranes such as the mouth, nose, eyes, rectum, udder, and genitalia. Table \ref{lsd_stages} shows the progression of LSD, detailing symptoms across stages from fever and lethargy in the early phase to the development of painful skin lesions. A growing body of research is leveraging ML to predict the occurrence of infectious diseases in animal populations. Specifically, ML methods have been developed to predict the onset of LSD infections. Researchers use ML techniques to analyze image data from infected animals, with these methods proving effective in predicting LSD and other viral diseases in test datasets ~\cite{girma2021identify}. Ananda et al. \cite{ananda2025machine} proposes using deep CNN feature extractors  i.e. VGG‑16, VGG‑19, Inception‑v3, combined with machine‑learning classifiers for automated LSD detection, reporting improved feature extraction and better performance versus classical methods.

Neha et al. \cite{ujjwal2022exploiting} tested the performance of the Random Forest model for detecting LSD in various cattle. They also compared the performance of other machine learning techniques. Raj et al. \cite{raj2023automated} proposed a hybrid model based on VGG-19 and ResNet-50 deep learning architectures to accurately detect LSD conditions. They used principal component analysis to decrease the dimensionality of the feature vector. Liang et al. \cite{liang2024lumpy} utilized machine learning techniques to predict outbreaks of African Swine Fever by analyzing global bioclimatic data. While other methods achieved an accuracy of 80.4\%, the Random Forest technique proved more accurate than the SVM when applied to a dataset of key climate variables, with an accuracy of 76.02\%. Niu et al. \cite{niu2021prediction} employed machine learning methods to predict outbreaks of Peste des Petits Ruminants using bioclimatic variables and altitude data. Their model's precision ranged from 47\% to 96\%, with the Random Forest method showing superior performance when applied to data from three countries not included in the training dataset. 
Figure. \ref{abstract} summarizes the role of our proposed LUMPNet in Precision livestock farming and early disease detection.

\begin{figure}
\centering
\includegraphics[height = 14.5cm, width = 8.6cm]{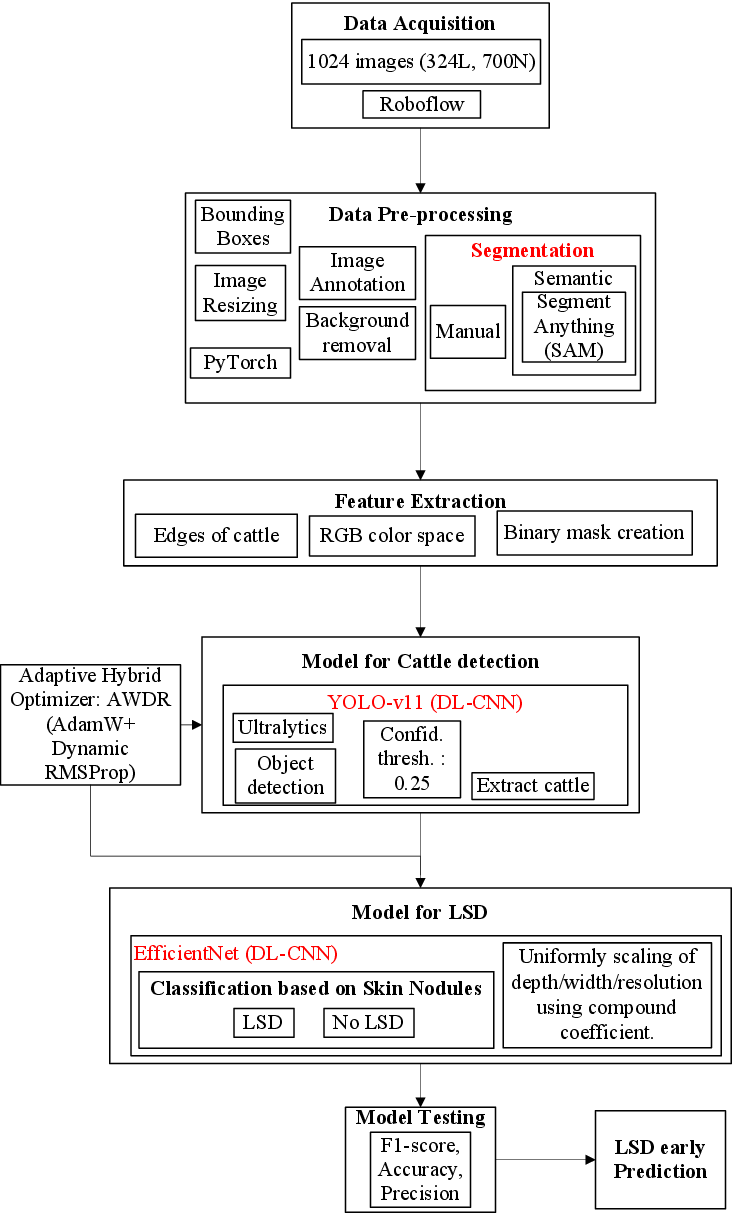}
\caption{Proposed system architecture  \label{flowchart}}
 \label{fig:architecture}
\end{figure}

\subsection{Novelty and contributions}
A number of deep learning (DL)-based models have been proposed in the literature for early detection of LSD-affected cattle \cite{muhammad2024lumpy}. However, most of these models do not generalize across diverse environments and data scarcity, leading to suboptimal performance. Additionally, the traditional feature extraction methods adopted by the recent literature fail to effectively classify LSD cases, particularly in the early stages when symptoms, such as skin nodules, are less pronounced. Moreover, most of these models take a long converge time and are not scalable, making them inefficient to be deployed in real-world livestock monitoring systems \cite{morgan2024exploring}. 

To address these issues, we propose a unified, fast‑converging, and scalable deep‑learning pipeline approach called LUMPNet, which integrates advanced object detection, classification, and intelligent optimization techniques to enhance the accuracy of early-stage LSD detection by focusing on the visual appearance of skin nodules. The modular design of LUMPNet enables it to support multi-domain adaptability, enabling re-training for new object categories or environmental conditions with minimal overhead. The novelty and cotributions are as follows. 
\begin{itemize}
    \item LUMPNet combines YOLOv11-based detection with an EfficientNet-based CNN classifier for accurate localization and precise classification of the affected animals. The EfficientNet-based CNN classifier integrates compound scaling in depth, width, and resolution to enhance accuracy.
    \item An Adaptive Hybrid optimizer (AWDR) is proposed, which combines the variance smoothing capability of RMSProp with the decoupled weight-decay regularization of AdamW through a dynamic blending coefficient. The optimizer enhances gradient stability, accelerates convergence, mitigates overfitting, and improves generalization on small or imbalanced datasets.
    \item Extensive evaluations are conducted to demonstrate that LUMPNet outperforms state-of-the-art baselines in terms of precision, recall, F1-score, and inference latency by around.
\end{itemize}

  Table \ref{tab:comparison1} summarizes the comparison of LUMPNet with the existing approaches. The rest of the paper is structured as follows. Section 2 presents the proposed model and methodology. Section 3 discusses the case study in which we compare the previous models with the EfficientNet baseline model. Section 4 explains the results, and Section 5 concludes the paper. 

\begin{figure}[t]
\centering
\includegraphics[height = 5.5cm, width = 8cm]{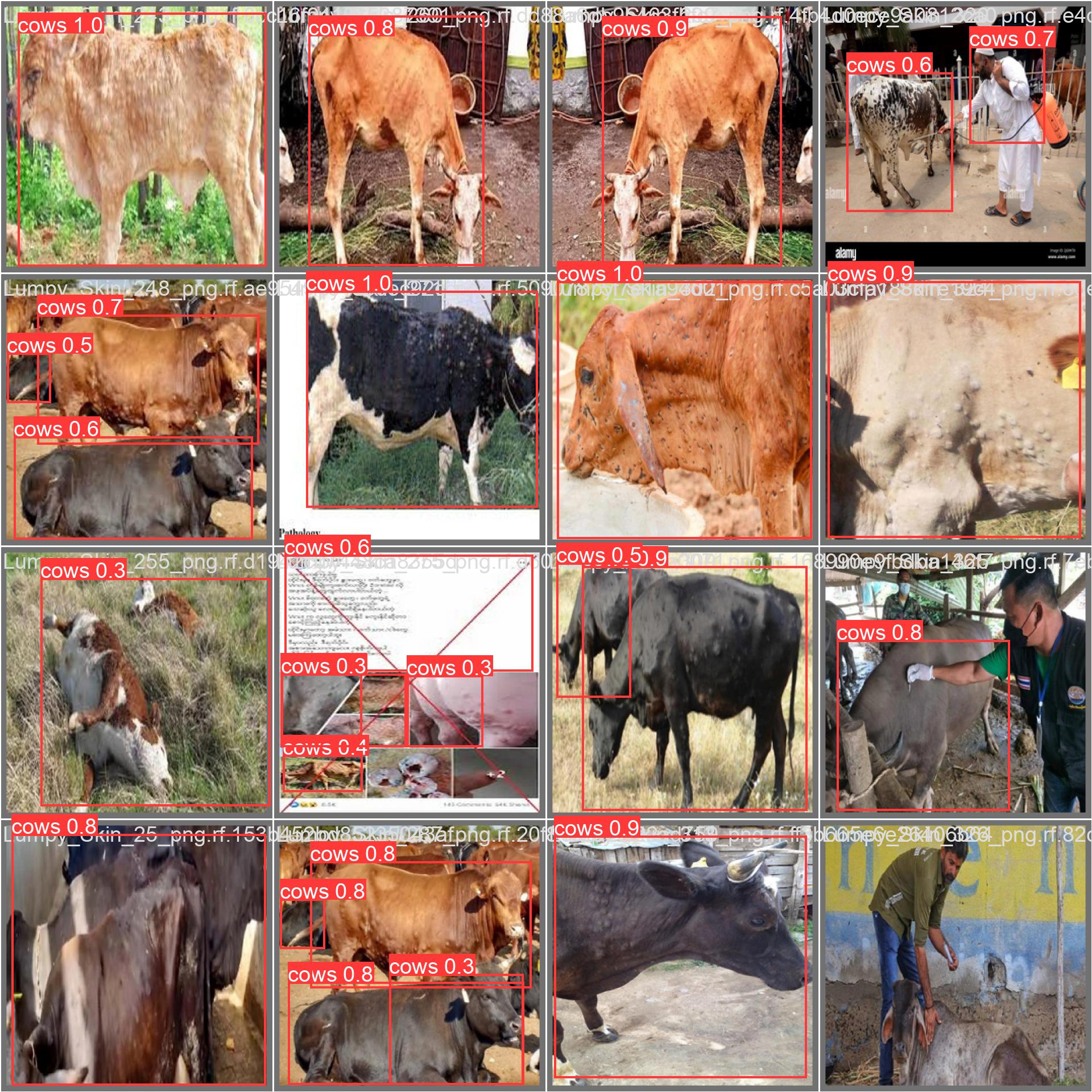}
\caption{YOLOv11 validation batch 1 predictions output in the proposed model\label{val_batch}}
\end{figure}

\section{Methodology}

LUMPNet detects LSD in cattle by integrating deep learning models i.e. YOLOv11 and EfficientNet, with novel AWDR optimizer. More precisely, the model uses a multifaceted approach encompassing object detection and semantic segmentation techniques. The process begins by identifying the cattle in the dataset \cite{lsd2024}. Both manual and automated segmentation is performed using the segment anything (SAM) package \cite{kirillov2023segment}. This results in separating the cattle's image from the colored background. Next, the main phase starts, where the symptoms of LSD, specifically skin nodules and lesions, are identified using YOLOv11 \cite{terven2023comprehensive}. The LSD-affected cattle is identified by finding the skin nodules or the affected skin. For classification, the model utilizes EfficientNet \cite{koonce2021efficientnet}. Figure \ref{flowchart} illustrates the system architecture, which comprises the following stages: dataset preparation, training a YOLOv11 model, segmentation, training an LUMPNet model for LSD classification, and evaluation.

In this paper, we use image dataset publicly available on Kaggle \cite{lsd2024}. The dataset contains 1024 cattle images, in which both healthy and LSD affected cattle images exist. In these images, the LSD affected cattle have a mild to extreme appearance of skin nodules.  The appearance of nodules characterizes the multiple stages of LSD. Among these images, there are 324 images of cattle with lumpy skin and an additional 700 images of healthy cattle. A key challenge was to develop an effective partitioning strategy that would support accurate model training and evaluation. To this end, 500 images of healthy cattle were allocated for training, while the remaining 200 were designated for testing. Similarly, 224 images of LSD-affected cattle were selected for training, and 100 images were reserved for testing.

\begin{table*}
\centering
\caption{Extracted Features for LSD Detection}
\begin{tabular}{|p{4cm}|p{10cm}|p{2cm}|}
\hline
\textbf{Feature Name} & \textbf{Description} & \textbf{Type} \\
\hline
Nodule Count & Total number of visible nodules detected on the skin surface & Quantitative \\
Nodule Size & Average area or diameter of individual nodules & Quantitative \\
Nodule Shape & Circularity, aspect ratio, and irregularity of nodules & Morphological \\
Edge Density & Density of edges (sharp transitions) around nodules & Structural \\
YOLOv11 Region Proposals & Bounding box features: position, size, confidence score & Learned Feature \\
EfficientNet Embeddings & Deep learned features extracted from intermediate convolutional layers & Learned Feature \\
\hline
\end{tabular}
\label{tab:lsd_features}
\end{table*}

\subsection{Image Pre-processing}
During image pre-processing, we perform both manual and automatic segmentation. 
Moreover, we mask the background color of the images with blue to maintain consistency, enhance the focus of the cattle's skin, and ultimately improve the accuracy of the model. This leads to the accurate detection of skin-related features. For automatic segmentation, we utilize SAM, which provides more accurate region detection. We improve model prediction by visualizing masks, points and bounding boxes through the utility functions. For object detection, we use ultralytics, NumPy, OpenCV, PyTorch and Matplotlib. Moreover, we integrate Roboflow with the system architecture and model training pipelines in order to achieve automated data management and efficient model training. We perform manual segmentation on the dataset to ensure higher quality and more precise data. This segmentation is important to effectively training the model and accurately evaluating its performance. 

\subsection{Feature Extraction}
The feature extraction stage operates on the cattle region obtained after pre-processing/segmentation and produces three complementary cues that are fed to the learning models as explained in Table \ref{tab:lsd_features}. Specifically, we compute (1) Edges of cattle: gradient-based edge maps to emphasize nodule boundaries and local shape irregularities; (2) RGB color space: the native RGB representation with per-channel statistics that preserve appearance and pigmentation differences associated with lesions; and (3) Binary mask creation: a foreground mask isolating the cattle silhouette to suppress background clutter and constrain subsequent analysis. Formally, for an input image $x$ and its segmentation $\hat{M}$, we derive $E=\text{edge}(x\odot\hat{M})$, $C=x_{\text{RGB}}$, and $B=\mathbb{1}(\hat{M}>0)$, and form a composite feature tensor $F=[E,\,C,\,B]$. During training, $F$ is paired with image-level labels (healthy vs.\ LSD-affected) to supervise the models; at test time, the same features guide prediction. The above-mentioned features provide complementary geometric and photometric evidence that improves nodule localization.

\subsection{Implementation of YOLOv11}
\begin{figure*}[t]
  \centering
  \includegraphics[trim=1cm 3.4cm 1cm 1cm, clip, width=0.82\textwidth]{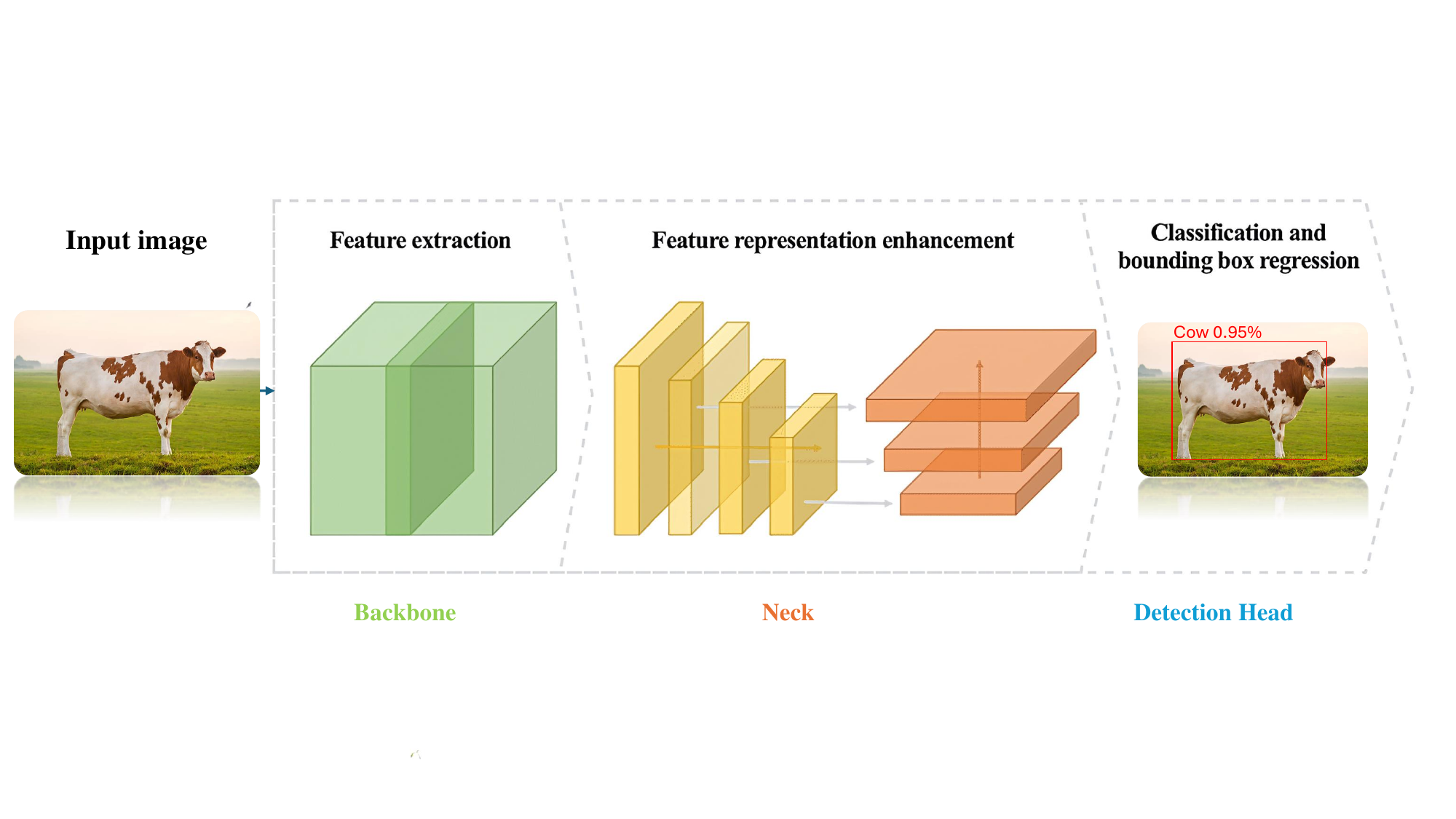}
  \caption{Architecture of YOLOv11 algorithm}
  \label{fig:yolo1}
\end{figure*}

YOLOv11 is a CNN-based framework for real-time object classification and detection \cite{terven2023comprehensive}. It introduces architectural enhancements that include Cross Stage Partial with kernel size 2 (C3k2) blocks for efficient feature reuse and Spatial Pyramid Pooling-Fast (SPPF) for multi-scale context. It strengthens spatial focus using Convolutional block with Parallel Spatial Attention (C2PSA). These characteristics enhance performance, particularly for detecting subtle LSD skin nodules in cattle. We implement pretrained weights via transfer learning and generate detection plots for qualitative evaluation. YOLOv11 partitions an input image into multiple grid cells. Each cell predicts bounding boxes with associated confidence scores. Figure \ref{fig:yolo1} illustrates the complete architectural design of the YOLOv11 algorithm. 

During bounding box regression, the regression model predicts these parameters via the Distribution Focal Loss (DFL) mechanism, permitting localization of LSD lesions on livestock pores and skin. YOLOv11 replaces C2f with the C3k2 block, so it provides proficient implementation of the Cross Stage Partial (CSP) Bottleneck. This architectural innovation helps gradient float via deep network layers, stopping vanishing gradient issues at the same time as keeping faster processing speeds. Residual connections permit the community to study complicated capabilities vital for subtle lesion detection. Intersection Over Union (IoU) quantifies the overlap amongst anticipated and floor fact bounding bins. It is mathematically described as
  \begin{equation}
  \text{IoU} = \frac{\text{Area of Overlap}}{\text{Area of Union}} = \frac{|B_{\text{pred}} \cap B_{\text{gt}}|}{|B_{\text{pred}} \cup B_{\text{gt}}|} .
  \end{equation}

YOLOv11 utilizes Complete IoU (CIoU) loss that promises high-precision lesion localization.

\subsubsection{Loss function formulation}
The loss function of YOLOv11 combines Complete Intersection over Union (CIoU) loss, Distribution Focal Loss (DFL), and Binary Cross-Entropy (BCE) loss. The total loss is expressed as a weighted sum
\begin{equation}
\mathcal{L}_{\text{total}} = \lambda_{\text{box}} \mathcal{L}_{\text{CIoU}} + \lambda_{\text{dfl}} \mathcal{L}_{\text{DFL}} + \lambda_{\text{cls}} \mathcal{L}_{\text{BCE}} ,
\end{equation}
where $\lambda_{\text{box}} = 7.5$, $\lambda_{\text{dfl}} = 0.5$, and $\lambda_{\text{cls}} = 1.5$ are the default weighting factors that balance the relative importance of each loss component. The CIoU loss incorporates three geometric factors: overlap area, center point distance, and aspect ratio consistency. It is formulated as \cite{YOLOmath},
\begin{equation}
\mathcal{L}_{\text{CIoU}} = 1 - \text{IoU} + \frac{\rho^2(b, b^{gt})}{c^2} + \alpha v ,
\end{equation}
where $\rho^2(b, b^{gt})$ represents the Euclidean distance between the center points of the predicted box $b$ and ground truth box $b^{gt}$, $c$ denotes the diagonal length of the smallest enclosing box covering both boxes, and $v$ measures the consistency of aspect ratios. Distribution Focal Loss models bounding box boundaries as probability distributions over discrete bins. It is formulated as, 
\begin{equation}
\begin{split}
\mathcal{L}_{\mathrm{DFL}}(P, y) \;=\; &\; -\sum_{i=0}^{n-1} \Big[ \; (y_i^{+} - y)\,\log\big(p_i\big) \\
&\qquad\qquad\qquad\qquad\; +\; (y - y_i^{-})\,\log\big(p_{i+1}\big) \; \Big] ,
\end{split}
\end{equation}
where $y$ represents the ground truth coordinate, $y_i^-$ and $y_i^+$ are the floor and ceiling integer bins surrounding $y$, and $p_i$ denotes the predicted probability for bin $i$. DFL also captures uncertainty in bounding box localization, particularly beneficial for detecting LSD lesions with ambiguous or blurry boundaries. Binary Cross-Entropy (BCE) loss optimizes class predictions for each detected object. For the binary LSD classification task (healthy vs. infected), the BCE loss is computed as
\begin{equation}
\mathcal{L}_{\text{BCE}} = -\frac{1}{N} \sum_{i=1}^{N} \left[ y_i \log(\hat{p}_i) + (1 - y_i) \log(1 - \hat{p}_i) \right] , 
\end{equation}
where $N$ represents the total number of predictions, $y_i \in \{0, 1\}$ denotes the ground truth label, and $\hat{p}_i$ is the predicted probability for class presence. 

\begin{figure*}[ht]
    \centering
    \includegraphics[
        width=1\textwidth, 
        clip 
    ]{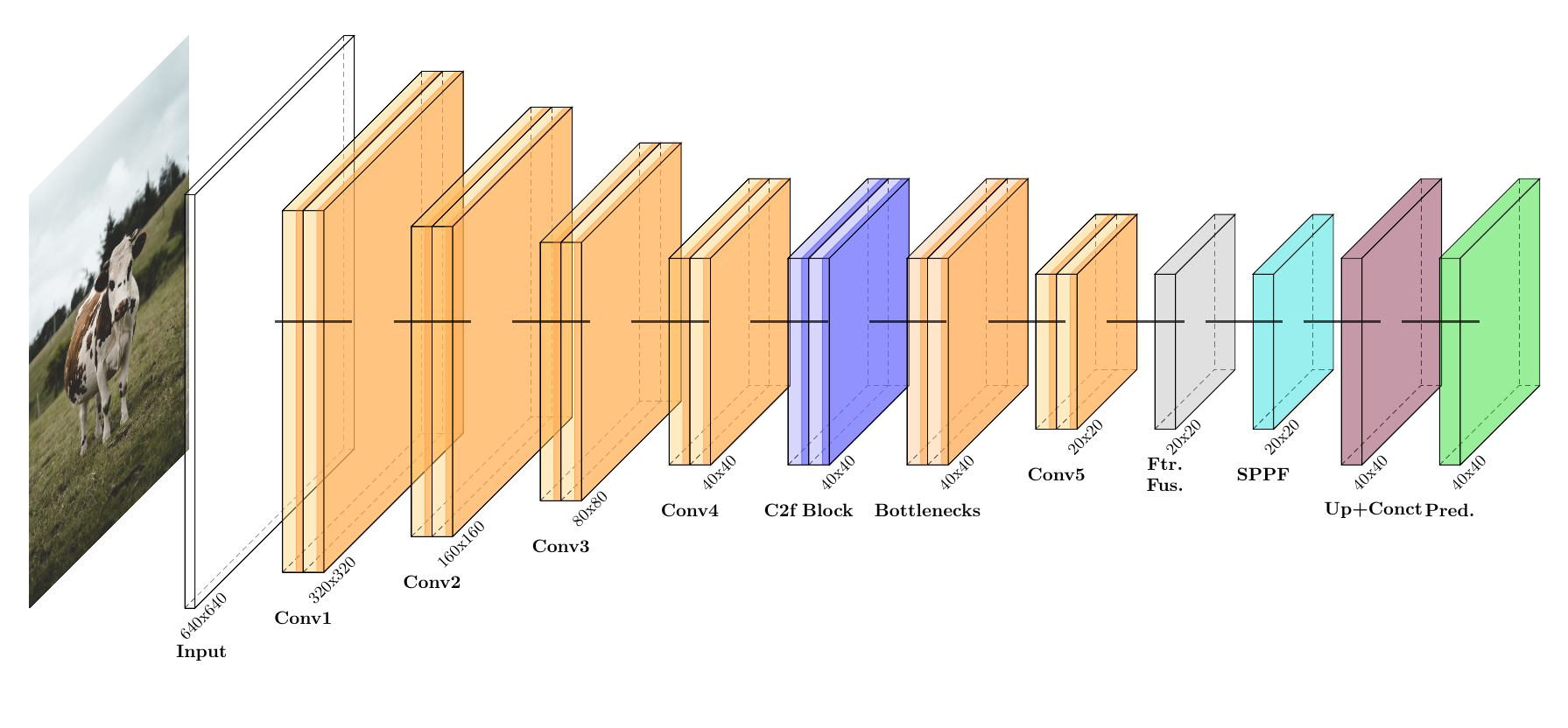}
    \caption{YOLO Model Layer Dimensions}
    \label{tab:yolostagesall}
\end{figure*}

As demonstrated in Figure. \ref{tab:yolostagesall}, we apply $3\times3$ convolutions with stride $(2,2)$ to downsample while increasing feature depth; each block uses batch normalization and SiLU activations to stabilize training and enable nonlinear pattern learning. Channel widths scale from 32 to 64, 128, 256, 512, and 1024, capturing progressively deeper features for robust detection of LSD skin nodules. 

\subsubsection{Architecture of YOLOv11}
Our model of YOLOv11 contains a Spatial Pyramid Pooling-Fast (SPPF) block. This block aggregates features from several receptive field sizes through sequential max-pooling operations, mathematically expressed as, 
\begin{equation}
\begin{split}
\mathrm{SPPF}(x) \;=\; &\;\mathrm{Conv}_{1\times 1}\Big( 
\mathrm{Concat}\Big[\, x,\; \mathcal{M}_k(x),\\
&\qquad\; \mathcal{M}_k\big(\mathcal{M}_k(x)\big),\;
\mathcal{M}_k\big(\mathcal{M}_k(\mathcal{M}_k(x))\big) \,\Big] \Big) ,
\end{split}
\end{equation}
where $\mathcal{M}_k$ represents max-pooling with kernel size $k$ (typically $k=5$) and stride 1, and $\text{Concat}$ denotes channel-wise concatenation. The C2PSA block permits the network to selectively cognizance on critical spatial areas in the feature maps. It additionally computes attention weights primarily based on spatial correlations
\begin{equation}
\mathbf{A}_{\text{spatial}} = \sigma\left( \text{Conv}_{7 \times 7}\left( \text{Concat}[\text{MaxPool}(\mathbf{F}), \text{AvgPool}(\mathbf{F})] \right) \right), 
\end{equation}
where $\mathbf{F}$ represents the input feature map, $\sigma$ denotes the sigmoid activation function, and the attention-refined features are obtained through
\begin{equation}
\mathbf{F}_{\text{refined}} = \mathbf{A}_{\text{spatial}} \odot \mathbf{F} ,
\end{equation}

This attention mechanism allows YOLOv11 to concentrate computational resources on potentially lesion-containing regions while suppressing irrelevant background features, dramatically improving detection accuracy for small or partially occluded LSD nodules, a common challenge in field conditions where cattle have varying coat colors, lighting conditions, and occlusions from environmental factors. The model architecture continues to grow deeper with more convolutional blocks and feature fusion operations through the neck component, enabling the extraction of both low-level and high-level features. Lower-level features extracted in earlier layers focus on fine details, including edges, texture patterns, color variations, and subtle skin surface changes characteristic of early-stage LSD lesions. 

During the detection phase, the neck component employs Path Aggregation Network (PANet) principles with upsampling and concatenation techniques to merge features from different network layers. The neck processes features at multiple scales: P3 (high-resolution, $80 \times 80$), P4 (medium-resolution, $40 \times 40$), and P5 (low-resolution, $20 \times 20$). Feature fusion is achieved through
\begin{equation}
\mathbf{F}_{\text{fused}}^{(i)} = \text{C3k2}\left( \text{Concat}\left[ \mathbf{F}_{\text{backbone}}^{(i)}, \text{Upsample}(\mathbf{F}_{\text{fused}}^{(i+1)}) \right] \right)
\end{equation}
for the top-down pathway, and
\begin{equation}
\mathbf{F}_{\text{out}}^{(i)} = \text{C3k2}\left( \text{Concat}\left[ \mathbf{F}_{\text{fused}}^{(i)}, \text{Downsample}(\mathbf{F}_{\text{out}}^{(i-1)}) \right] \right)
\end{equation}
for the bottom-up pathway. This bidirectional feature fusion enables the model to generate multi-scale predictions that can detect both small isolated nodules on the neck and shoulders and larger lesion complexes on the body trunk and limbs.

The final detection head generates predictions at three different scales, outputting for each anchor point: (1) class probabilities $p_{\text{cls}} \in [0,1]^C$ for $C$ classes, (2) bounding box coordinates $(x, y, w, h)$ modeled as probability distributions through DFL, and (3) objectness scores $p_{\text{obj}} \in [0,1]$ indicating detection confidence. The detection head employs decoupled prediction branches where classification and localization are processed through separate convolutional pathways, reducing the task conflict and improving overall performance. For each prediction scale, the head uses small $1 \times 1$ convolutions to ensure precise predictions at each spatial location, 
\begin{equation}
\begin{aligned}
p_{\text{cls}} &= \sigma(\text{Conv}_{1 \times 1}^{\text{cls}}(\mathbf{F}_{\text{out}})) \\
\mathbf{d}_{\text{box}} &= \text{Softmax}(\text{Conv}_{1 \times 1}^{\text{box}}(\mathbf{F}_{\text{out}})) \\
p_{\text{obj}} &= \sigma(\text{Conv}_{1 \times 1}^{\text{obj}}(\mathbf{F}_{\text{out}})) ,
\end{aligned}
\end{equation}
where $\mathbf{d}_{\text{box}}$ represents the distribution over discrete coordinate bins for DFL. 

Overall, the YOLOv11 model processes images of size $640 \times 640 \times 3$. The spatial dimensions progressively reduce due to strided convolutions and pooling operations: $640 \times 640 \rightarrow 320 \times 320 \rightarrow 160 \times 160 \rightarrow 80 \times 80 \rightarrow 40 \times 40 \rightarrow 20 \times 20$, while the depth of feature maps increases systematically: $3 \rightarrow 32 \rightarrow 64 \rightarrow 128 \rightarrow 256 \rightarrow 512 \rightarrow 1024$ channels. During the stages, the model captures complex hierarchical features and combines multi-scale feature maps from three resolution levels through the neck's feature fusion operations. These fused results are processed by the detection head to generate accurate predictions for detecting LSD in cattle.

\begin{figure*}[t]
  \centering
  \includegraphics[height = 5cm, width = 15cm]{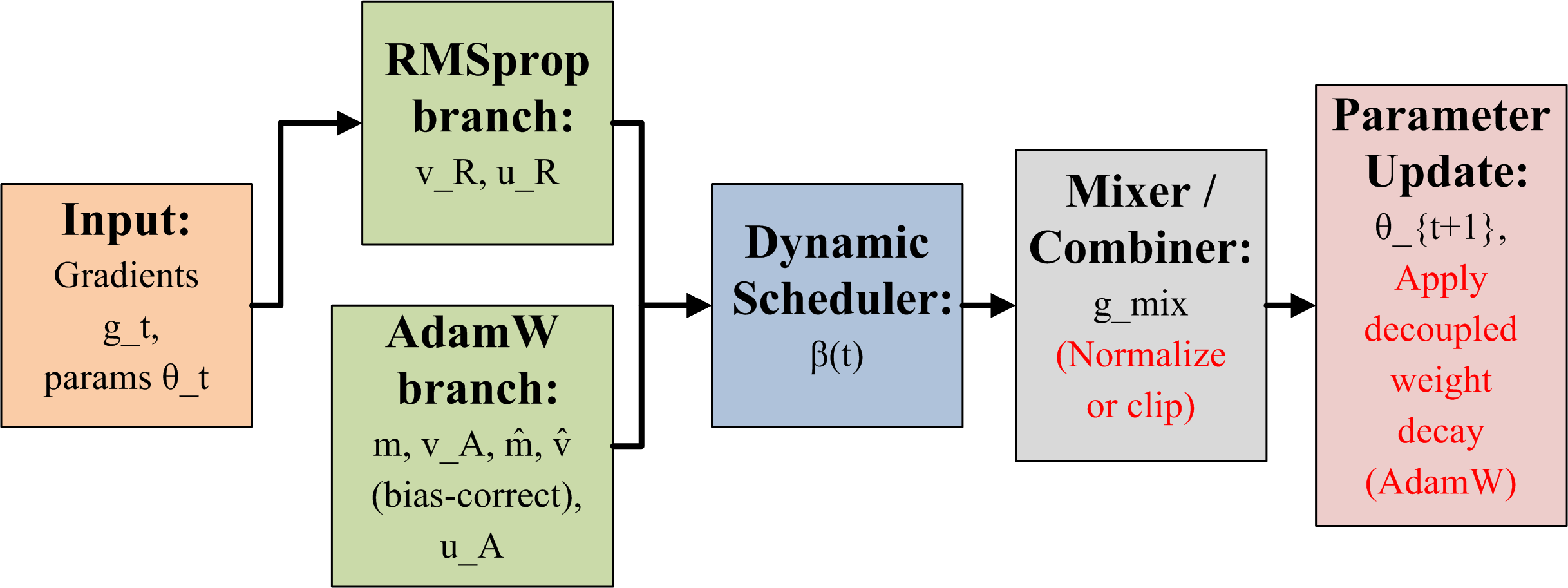}
  \caption{Architecture of AWDR optimizer}
  \label{fig:opt}
\end{figure*}

\subsection{Adaptive Hybrid Optimizer: AWDR}
We devise an AWDR optimizer that combines the weight decay regularization capabilities of AdamW with the adaptive variance smoothing nature of RMSprop. The proposed fusion model not only enables stable gradient updates during the early stages of training, moreover, preserves better generalization in later epochs. The optimizer implements a dynamic transition schedule. In this scheduling mechanism, time-dependent coefficient $\beta(t)$ controls the gradual shift between RMSprop-like behavior towards AdamW-dominant behavior. The dynamic transition is formulated as,
\begin{equation}
\beta(t) = \beta_{0} \cdot \left(1 - \frac{t}{T}\right), 
\end{equation}
where $t$ denotes the current epoch, $T$ is the maximum number of epochs, and $\beta_{0}$ is an initialization factor. The AWDR update rule is then expressed as,
\begin{equation}
g_{t} = \beta(t) \cdot RMSprop(g_{t}) + \left(1 - \beta(t)\right) \cdot AdamW(g_{t}).
\end{equation}
This optimizer results in more stable convergence and reduced overfitting. Therefore, RMSprop contributes stronger gradient smoothing during early high-variance updates, while AdamW progressively regulates weight decay in the later optimization stages. It also provides three main advantages over traditional singular optimizer, i.e.,
\begin{itemize}
    \item Improved training stability across long training cycles
    \item Faster convergence with reduced gradient oscillations
    \item Better generalization performance due to AdamW-based weight decay enforcement
\end{itemize}
Consequently, AWDR significantly enhances the learning capability of our proposed hybrid architecture for Lumpy Skin Disease detection, enabling higher robustness and improved classification accuracy. Figure. \ref{fig:opt} explains the architecture of the proposed optimizer.
\begin{figure}
\centering
\includegraphics[height = 13cm, width = 10cm, trim=1cm 13cm 10cm 1cm, clip]{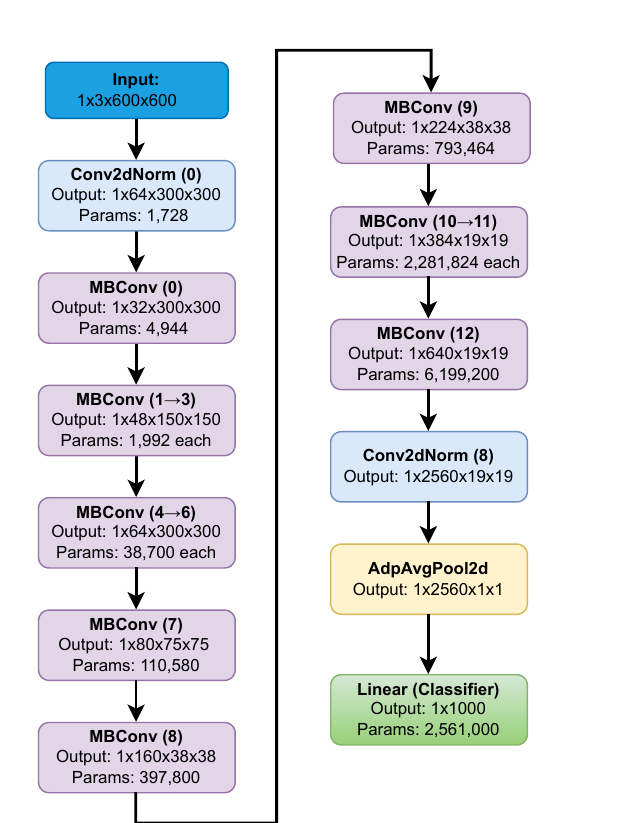}
\caption{EfficientNet Model Layer Dimensions \label{eff_str}}
\end{figure}

\begin{figure}[t]
\centering
\includegraphics[height = 5cm, width = 6.8cm]{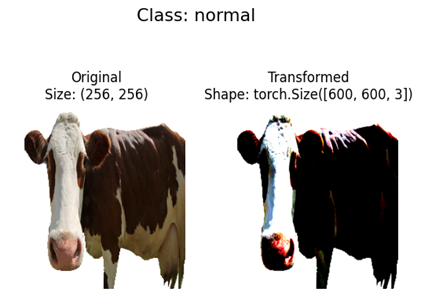}
\caption{Images showing the accuracy of EfficientNet in transforming Class: Normal  cattle\label{image_normal}}
\end{figure}

\subsection{EfficientNet-based Classification}
EfficientNet is used as the main component for LSD detection in our model as demonstrated in Figure. \ref{eff_str}. It is a CNN architecture equipped with the concept of uniformly scaling a model, called \textit{compound scaling}, a crucial advancement in deep learning. The compound scaling strategy is used to solve the trade-off among accuracy, efficiency, and size during CNN scaling. In this strategy, a compound coefficient, denoted as $\phi$, is applied to scale the network's width, depth, and resolution in a balanced and systematic manner, as given below
\begin{equation}
\begin{aligned}
\text{depth:} \quad & d = \alpha \phi \\
\text{width:} \quad & w = \beta \phi \\
\text{resolution:} \quad & r = \gamma \phi \\
\text{s.t.} \quad & \alpha \cdot \beta^2 \cdot \gamma^2 \approx 2 \\
& \alpha \geq 1 \\
& \beta \geq 1 \\
& \gamma \geq 1 .
\end{aligned}
\end{equation}
Here, $\alpha$, $\beta$, and $\gamma$ are constants that are determined through a small grid search. Conceptually, $\phi$ is a coefficient particular to the consumer to manipulate the general scaling of the model, indicating how many extra assets should be allotted for scaling. Meanwhile, $\alpha$, $\beta$, and $\gamma$ determine how those properties are disbursed throughout the network’s width, depth, and resolution, respectively. This model approaches anomalies in cattle pictures with the use of transfer learning, which allows the model to quickly adapt to the primary features of the dataset. Therefore, it serves as a pretrained version for anomaly detection. Moving in advance with anomaly detection of the use of EfficientNet, the proposed model involves a couple of key steps. These consist of pre-processing responsibilities, which comprise of resizing images to the preferred input proportions for the model and augmenting the dataset to develop model generalization. 

\begin{figure}[t]
\centering
\includegraphics[height = 5cm, width = 6.8cm]{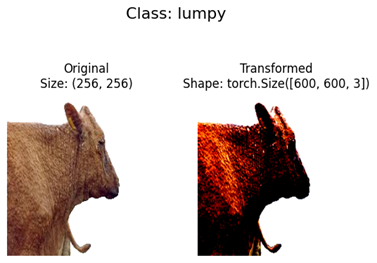}
\caption{Images showing the accuracy of EfficientNet in transforming Class: LSD affected cattle\label{image_lsd}}
\end{figure}

\begin{figure*}
\centering
\includegraphics[trim=1pt 1pt 1pt 8pt, clip, height = 9cm, width = 18cm]{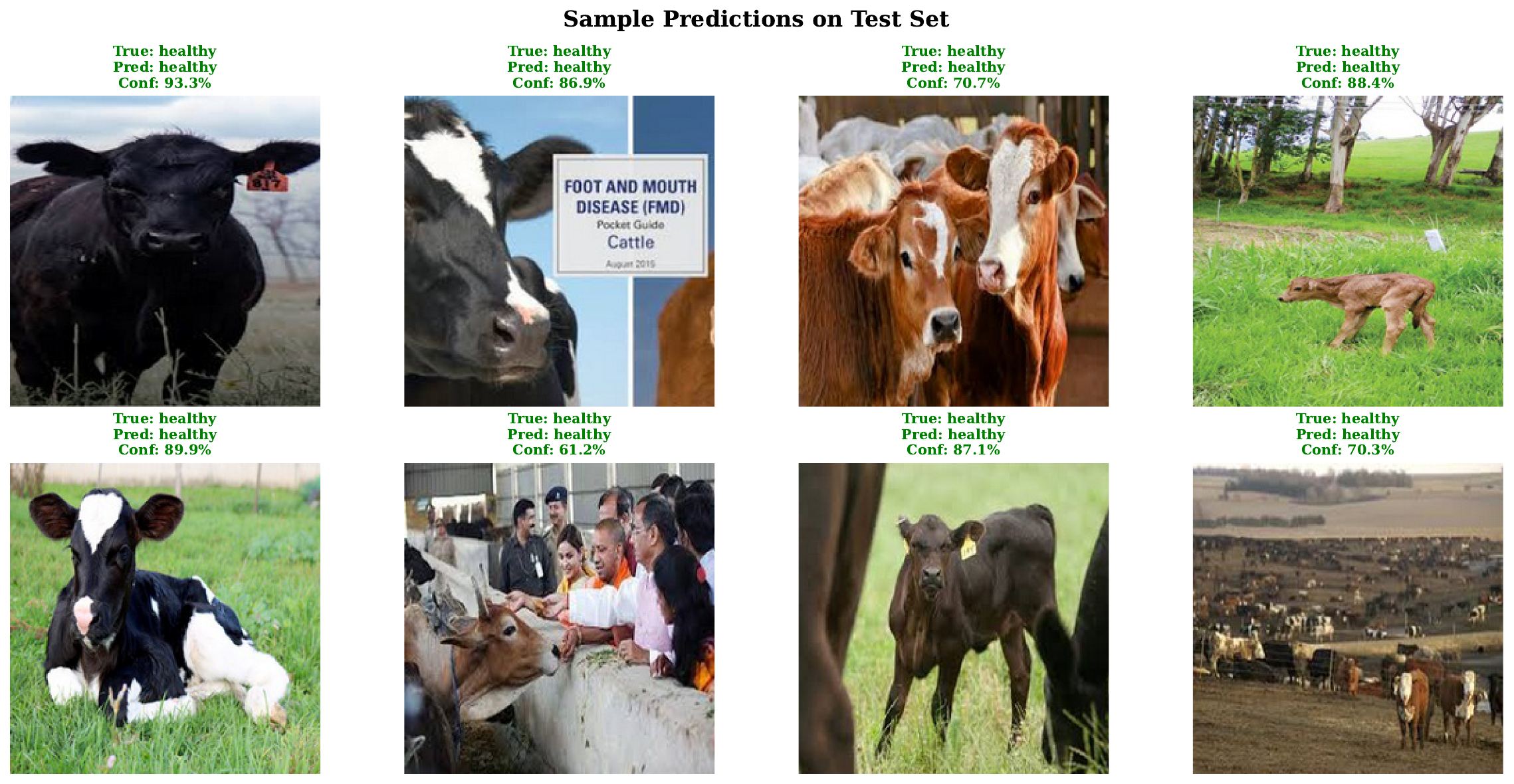}
\caption{Case study: Sample predictions on Test set\label{sample_pred_case}}
\end{figure*}

\begin{figure}
\centering
\includegraphics[height = 5cm, width = 7cm]{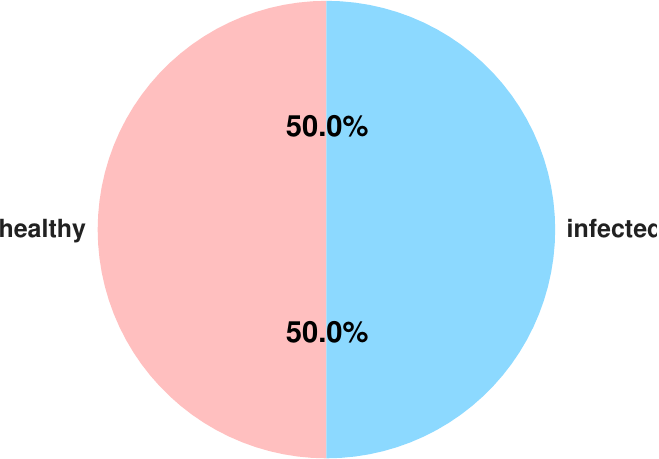}
\caption{Case study: Class distribution\label{class_distrb_case}}
\end{figure}

\begin{figure}
\centering
\includegraphics[height = 6cm, width = 8cm]{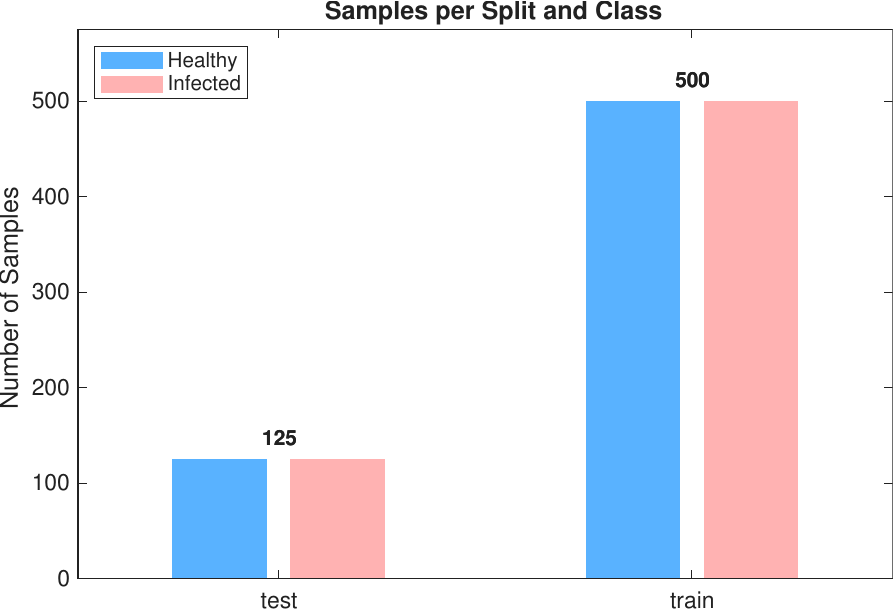}
\caption{Case study: Sample distribution\label{sample_distrib_case}}
\end{figure}

\begin{figure}
\centering
\includegraphics[height = 6cm, width = 8cm]{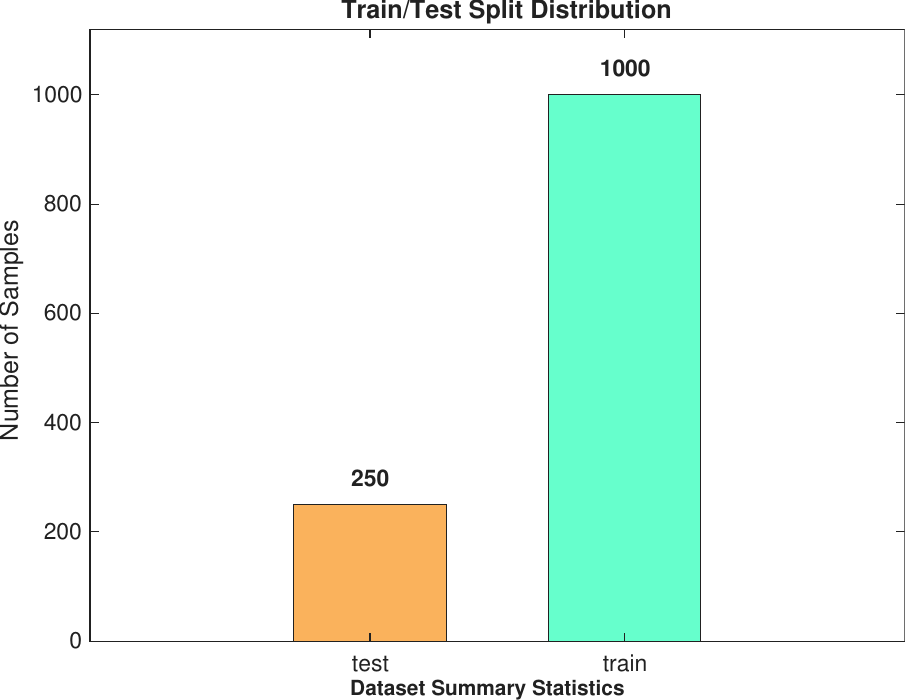}
\caption{Case study: Data Splitting\label{data_splt_case}}
\end{figure}

The EfficientNet model is selected as the base structure for transfer learning to know, making use of pretrained weights. We alter the model’s parameters on the training dataset to adapt it for LSD detection. To compare the model, we make predictions on the test dataset. Figures~\ref{image_normal} and~\ref{image_lsd} show the results of image classification using EfficientNet. EfficientNet model begins with a \texttt{Conv2dNormActivation} layer that performs a convolution using a 64-filter, 3$\times$3 kernel with stride adjustments, followed by batch normalization and the SiLU activation function. The model also uses MBConv blocks to  perform efficient feature extraction. These blocks reduce the spatial resolution while increasing the depth. 

The model utilizes an Adaptive Average Pooling layer, which aggregates the spatial features into a single vector. This pooled representation encapsulates the most relevant information from the input image and is ready to be passed into the final classification layers. The classifier part of the model consists of a Dropout layer, which helps prevent overfitting by randomly deactivating certain neurons during training, and a Linear layer that produces the final output. This output layer generates the classification results, which can be adjusted to handle specific tasks like distinguishing between LSD-affected cattle and healthy cattle. The entire EfficientNet model consists of multiple trainable parameters, making it capable of being fine-tuned during training for a given task, with no fixed or frozen layers.

EfficientNet begins with standard convolutional operations to extract spatial features from the input data. At each spatial location \((i, j)\), the output is obtained by applying a filter over a local neighborhood of the input feature map. The operation involves element-wise multiplication between the filter weights and the input region, followed by summation and the addition of a bias term
\begin{equation}
y_{i,j} = \sum_{m=-k}^{k} \sum_{n=-k}^{k} x_{i+m,j+n} \cdot w_{m,n} + b .
\end{equation}
Here, \(x_{i+m,j+n}\) represents the input at a neighboring position, \(w_{m,n}\) are the kernel weights, \(b\) is the bias term, and \(y_{i,j}\) is the resulting output at that position. The filter weights are learned during training to capture meaningful patterns in the data. To stabilize and speed up training, batch normalization is applied after convolutional operations. It normalizes the input features to have zero mean and unit variance, then scales and shifts the result using learnable parameters. The normalized output is given by
\begin{equation}
z_{i,j} = \gamma \left( \frac{x_{i,j} - \mu}{\sigma} \right) + \beta .
\end{equation}
In this equation, \(\mu\) and \(\sigma\) are the mean and standard deviation of the input features, while \(\gamma\) and \(\beta\) are learnable scaling and shifting parameters. 

EfficientNet uses the SiLU activation function that is defined as
\begin{equation}
\sigma(x) = x \cdot \frac{1}{1 + e^{-x}}.
\end{equation}
The core building blocks of EfficientNet are MBConv layers, which are based on MobileNetV2's inverted residuals and depthwise separable convolutions. First, a depthwise convolution is applied independently to each channel
\begin{equation}
y_{i,j,c} = \sum_{m=-k}^{k} \sum_{n=-k}^{k} x_{i+m,j+n,c} \cdot w_{m,n,c} .
\end{equation}
This operation focuses on spatial filtering within each channel. A pointwise convolution combines the outputs as follows:
\begin{equation}
y_{i,j} = \sum_{c} x_{i,j,c} \cdot w_{c} .
\end{equation}
At the end of the network, a fully connected (linear) layer maps the high-level features to class scores for classification tasks. For each output class \(k\), the score is computed as,
\begin{equation}
y_k = \sum_i x_i \cdot w_{i,k} + b_k .
\end{equation}
Here, \(x_i\) are the input features, \(w_{i,k}\) are the weights associated with class \(k\), and \(b_k\) is the corresponding bias. The final predicted class is the one with the highest score
\begin{equation}
\hat{y} = \arg\max(y_1, y_2, \dots, y_K) .
\end{equation}

\section{Case study}
\begin{figure}
\centering
\includegraphics[height = 5.5cm, width = 7.5cm]{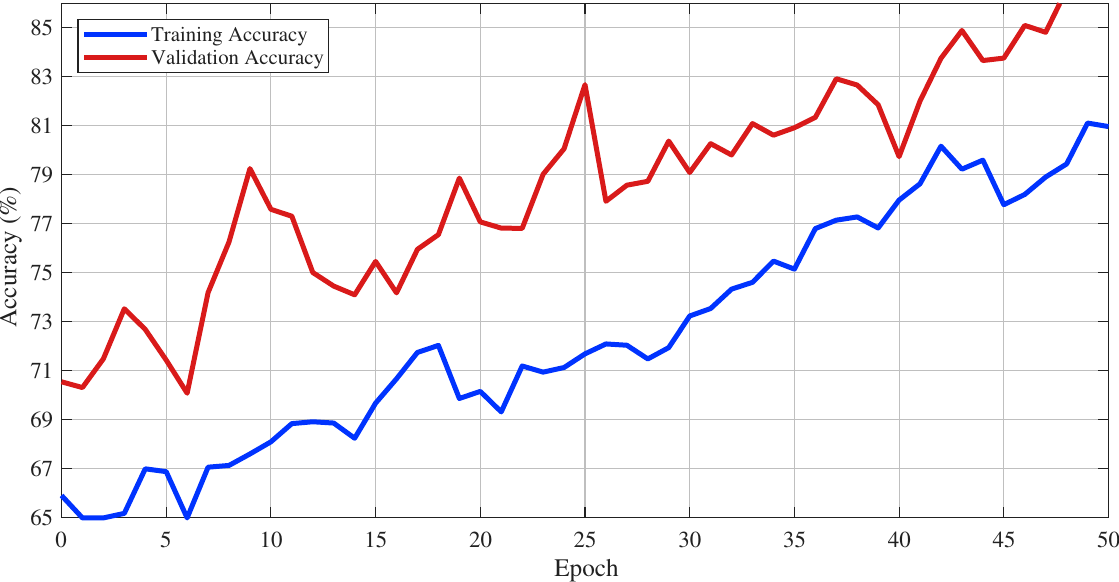}
\caption{Case study: Accuracy\label{accuracy_case}}
\end{figure}

\begin{figure}
\centering
\includegraphics[height = 5.5cm, width = 7.5cm]{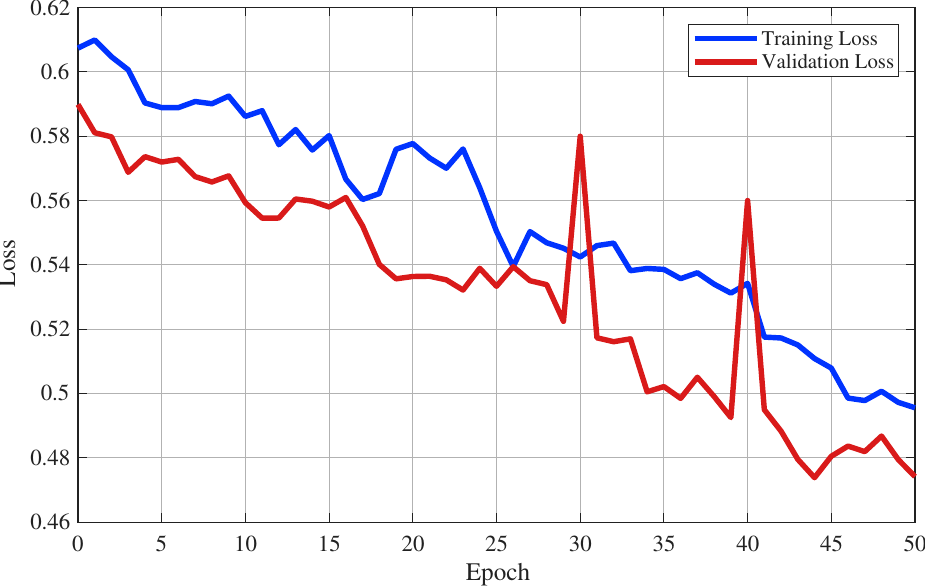}
\caption{Case study: Loss\label{loss_case}}
\end{figure}

We conducted a case study in which we implemented an optimized EfficientNet‑B0 model trained with the AdamW optimizer to compare its performance against our proposed LUMPNet. To evaluate the efficacy of the proposed image classification pipeline for the early detection of LSD in cattle, we analyze the learning dynamics of the EfficientNet‑B0 based model over 50 training epochs. Figure. \ref{sample_pred_case} - \ref{loss_case} illustrates the dataset details, and progression of training and validation accuracy, revealing several critical insights into the model's optimization and generalization behaviors. Both the training and validation accuracy metrics exhibit a robust positive correlation with the number of epochs, confirming that the model is effectively learning discriminative features from the augmented dataset. A noteworthy characteristic of the training trajectory is the persistent superiority of validation accuracy  over training accuracy. This phenomenon is consistent with the use of a heavy augmentation regimen, including random geometric transformations and color jitter, during the training phase. These augmentations introduce significant regularization by artificially increasing the difficulty of the training task, whereas the validation set remains relatively "clean," undergoing only resizing and normalization. Consequently, the model performs better on the simpler validation distribution than on the highly perturbed training data, a strong indicator that the model is not overfitting.

The volatility observed in the accuracy curves, characterized by sharp stochastic fluctuations (e.g., near epochs 8, 14, and 26), can be attributed to the aggressive learning rate modulation governed by the OneCycleLR scheduler. This scheduler, in conjunction with the AdamW optimizer, facilitates rapid exploration of the loss landscape, preventing the model from settling into sharp local minima early in the training process. By the final epoch, the validation accuracy surpasses 85\%, effectively outperforming the training accuracy which stabilizes near 80\%. 

The EfficientNet baseline attains 83.77\% accuracy with an F1‑score of 78.72\% on the validation split. The model yields an MCC of 0.6144 and a best observed accuracy of 85.71\%, indicating moderate-to-strong discriminative performance. Inference latency is  around 837.7 ms per image ($\sim$1.2 FPS), which limits real‑time use and suggests optimization. We conclude that the EfficientNet baseline delivers high efficacy on the LSD dataset, producing validation accuracies in the mid‑90s after stabilization. We treat this pipeline as a reliable baseline for comparing against our proposed LUMPNet hybrid system and we continue to refine it toward a deployable early‑warning tool for LSD detection in field settings.

\section{Results and Discussion}
The proposed model is tested on a high-performance machine equipped with 64 GB RAM, an NVIDIA RTX 4090 GPU, and a 13th Gen Intel Core i9 processor running Windows 11. For reproducibility and collaborative experiments, Google Colab is also utilized, offering seamless access to widely-used libraries, support for both GPU and TPU acceleration, and smooth integration with Google Drive for efficient data storage and sharing. The model is tested using the Lumpy skin image dataset, which is publicly available \cite{lsd2024}. We use Compute Unified Device Architecture (CUDA) to enhance the computational efficiency of the model. It also increases the speed of the model-training process. Table \ref{tab:1} compares the performance of the proposed techniques with the previous models \cite{ujjwal2022exploiting}.

\begin{table*}[t]
  \centering
  \caption{Performance comparison between the proposed LUMPNet and various previous models.}
  \scalebox{1.22}{%
\begin{tabular}{|c|p{3cm}|p{3cm}|c|c|c|}
\hline
\textbf{Model} & \textbf{Area Under Curve} & \textbf{Corr. Classif. Accuracy} & \textbf{F1 Score} & \textbf{Precision} & \textbf{Recall} \\
\hline
Random Forest & 0.995 & 0.977 & 0.977 & 0.977 & 0.977 \\
\hline
Neural Network & 0.992 & 0.962 & 0.962 & 0.962 & 0.962 \\
\hline
Adaboost & 0.992 & 0.972 & 0.972 & 0.972 & 0.972 \\
\hline
K-Nearest Neighbors & 0.983 & 0.961 & 0.961 & 0.962 & 0.961 \\
\hline
Decision Tree & 0.962 & 0.962 & 0.962 & 0.963 & 0.962 \\
\hline
Naive Bayes & 0.945 & 0.883 & 0.883 & 0.883 & 0.883 \\
\hline
Support Vector Machine & 0.482 & 0.432 & 0.379 & 0.396 & 0.432 \\
\hline
\textbf{LUMPNet} & \textbf{0.9968} & \textbf{0.9968} & \textbf{0.99} & \textbf{0.99} & \textbf{0.99} \\
\hline
\end{tabular}  }
\label{tab:1}
\end{table*}

\begin{table}[ht]
\centering
\caption{Comparison with similar studies.}
\setlength{\arrayrulewidth}{0.6pt} 
\setlength{\tabcolsep}{8pt}       
\renewcommand{\arraystretch}{1.4} 

\begin{tabular}{|p{6cm}|c|}
\hline
\textbf{Model} & \textbf{Accuracy} \\
\hline
Ensemble Method & 92\% \\
\hline
Extreme learning machine (ELM) & 90\% \\
\hline
DenseNet121 & 89\% \\
\hline
CNN-Model & 84\% \\
\hline
RMSProp+MobileNetV2 & 95\% \\
\hline
\textbf{LUMPNet} & \textbf{99}\% \\
\hline
\end{tabular}

\label{tab:comparison_2}
\end{table}

\begin{table}[ht]
\centering
\caption{Dataset Summary}

\setlength{\arrayrulewidth}{0.9pt}  
\setlength{\tabcolsep}{8pt}         
\renewcommand{\arraystretch}{1.4}   

\begin{tabular}{|p{6cm}|c|}
\hline
\textbf{Metric} & \textbf{Value} \\ \hline
Total Samples & \textbf{1250} \\ \hline
Training Samples & \textbf{1000} \\ \hline
Testing Samples & \textbf{250} \\ \hline
Healthy Samples & \textbf{625} \\ \hline
Infected Samples & \textbf{625} \\ \hline
Class Balance Ratio & \textbf{1:1.00} \\ \hline
\end{tabular}

\end{table}

\textit{Precision} is defined as the ratio of true positive predictions (\(TP\)) to the total number of positive predictions (\(TP + FP\)), where \(FP\) denotes false positives. In other words, precision evaluates the proportion of correct positive predictions made by the model.
\begin{equation}
    \text{Precision} = \frac{TP}{TP + FP} .
\end{equation}
Here, \(TP\) stands for true positives, and \(FP\) represents false positives. 
On the other hand, \textit{Recall} quantifies the model's ability to correctly identify true positives (\(TP\)) from all actual positive instances (\(TP + FN\)), where \(FN\) represents false negatives.
\begin{equation}
    \text{Recall} = \frac{TP}{TP + FN}
\end{equation}
Average Precision (AP) is calculated as the weighted mean of precision values across different recall thresholds. Alternatively, it can be interpreted as the proportion of true positives (\(TP\)) out of all detections (\(TP + FP\)) made by the model.
\begin{equation}
    \text{Average Precision} = \frac{TP}{TP + FP}
\end{equation}

Table \ref{tab:1} compares the performance of various ML and DL models with the proposed deep learning technique for LSD detection and classification. The comparison is based on five key performance metrics, i.e., Area Under the Curve (AUC), Classification Accuracy (CA), F1-score, Precision, and Recall. The K-Nearest Neighbors (KNN) model exhibits good performance across all metrics, achieving an AUC of 0.983, a classification accuracy of 0.961, and an F1-score of 0.961. Similarly, the Decision Tree model achieves an AUC of 0.962, with an identical accuracy, F1-score, precision, and recall of 0.962. The Random Forest model outperforms KNN and Decision Tree models, demonstrating an AUC of 0.995 and accuracy of 0.977, indicating its effectiveness in LSD detection. However, Support Vector Machine (SVM) significantly underperforms, with the lowest AUC (0.482), classification accuracy (0.432), and F1-score (0.379), suggesting its inefficiency in handling the dataset used for LSD classification. This could be attributed to the complexity of feature extraction and classification in LSD cases, where DL-based feature extraction may be more effective. The Neural Network model achieves an AUC of 0.992, an accuracy of 0.962, and an F1-score of 0.962, showing improved robustness compared to conventional ML models. Naïve Bayes, on the other hand, performs slightly worse, with an AUC of 0.945 and accuracy of 0.883, which may be due to its assumption of feature independence that does not align well with LSD features. The Adaboost model performs competitively, with an AUC of 0.972 and an accuracy of 0.972, demonstrating its ability to boost weak classifiers' performance. However, despite its good performance, it does not outperform the proposed model.

\begin{table}[ht]
\centering
\caption{Case study - Model Evaluation: Test dataset}
\label{tab:metrics}

\setlength{\arrayrulewidth}{0.9pt}  
\setlength{\tabcolsep}{8pt}         
\renewcommand{\arraystretch}{1.4}   

\begin{tabular}{|p{6cm}|c|}
\hline
\textbf{Metric} & \textbf{Value} \\ \hline
Test Accuracy (\%) & 78.57 \\ \hline
F1-Score Macro (\%) & 72.36 \\ \hline
Precision Macro (\%) & 77.25 \\ \hline
Recall Macro (\%) & 70.68 \\ \hline
Average Inference Time (ms) & 814.67 \\ \hline
Frames Per Second (FPS) & 1.2 \\ \hline
Model Parameters & 4,887,332 \\ \hline
Number of Classes & 2 \\ \hline
Total Test Samples & 160 \\ \hline
\end{tabular}

\end{table}

This improved performance may be attributed to DL's ability to extract rich, hierarchical information from photos of cattle impacted by LSD. The results presented in the table clearly demonstrate that our DL-based techniques perform better than traditional ML models for early LSD detection. The robustness of the suggested method in successfully detecting and categorizing LSD cases is implied by its high accuracy (99\%) and exceptional AUC, F1-score, precision, and recall. 

While the recall for healthy cattle is 1.0, indicating the model's ability to catch all non-inflamed instances without any false negatives, the recall for inflamed livestock is 0.98, indicating that the model successfully recognizes 0.98\% of inflamed cases, but it misses some. The model's overall best performance is highlighted by the F1 score, which strikes a compromise between precision and recall, being 1.0 for healthy cattle and 0.99 for inflamed farm animals. 

\begin{figure}
\centering
\includegraphics[height = 5.5cm, width = 7cm, trim=0cm 0cm 0cm 0.8cm, clip]{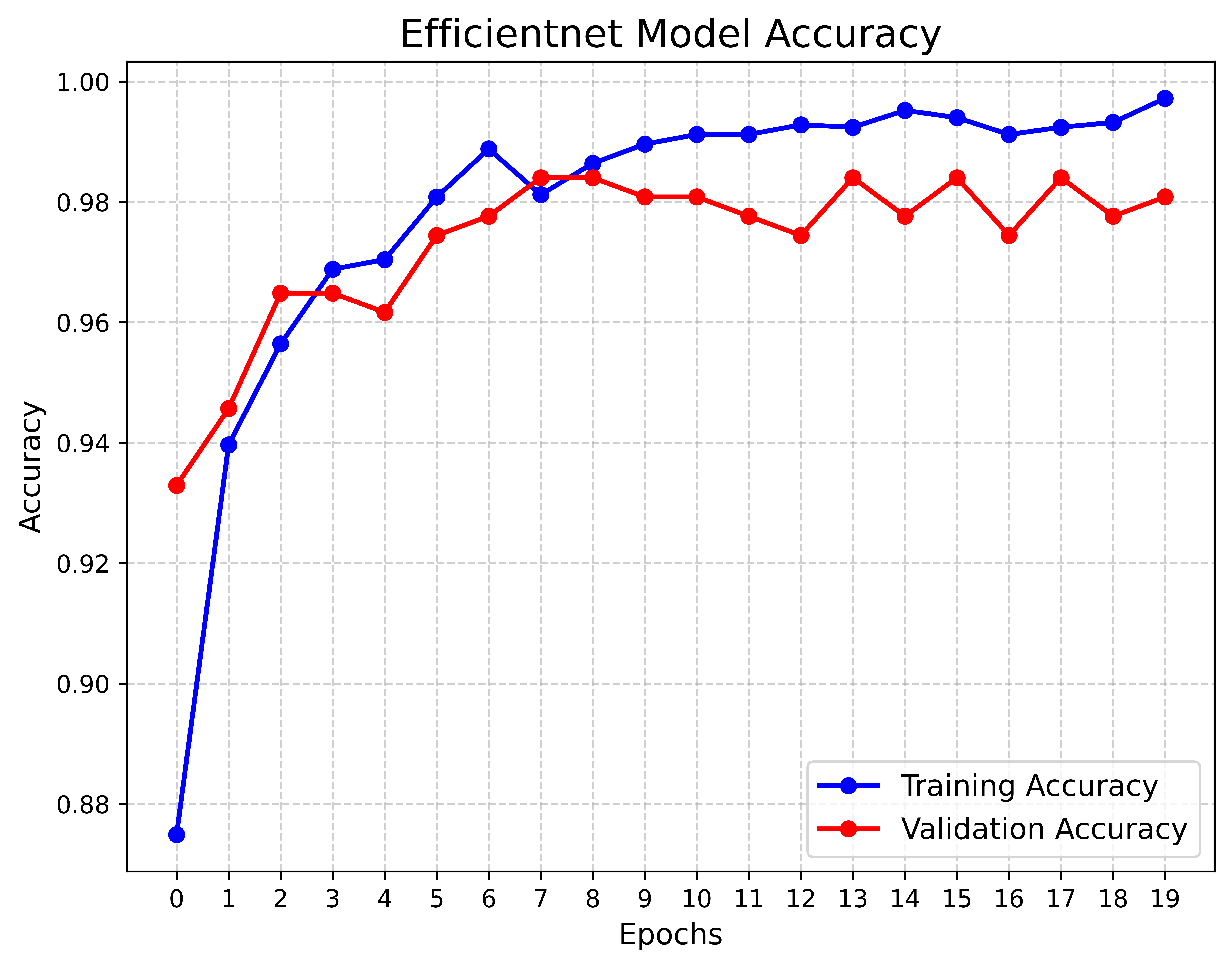}
\caption{LUMPNet Model Accuracy \label{eff_acc}}
\end{figure}

\begin{figure}
\centering
\includegraphics[height = 7cm, width = 8.5cm, trim=0cm 0cm 0cm 0.8cm, clip]{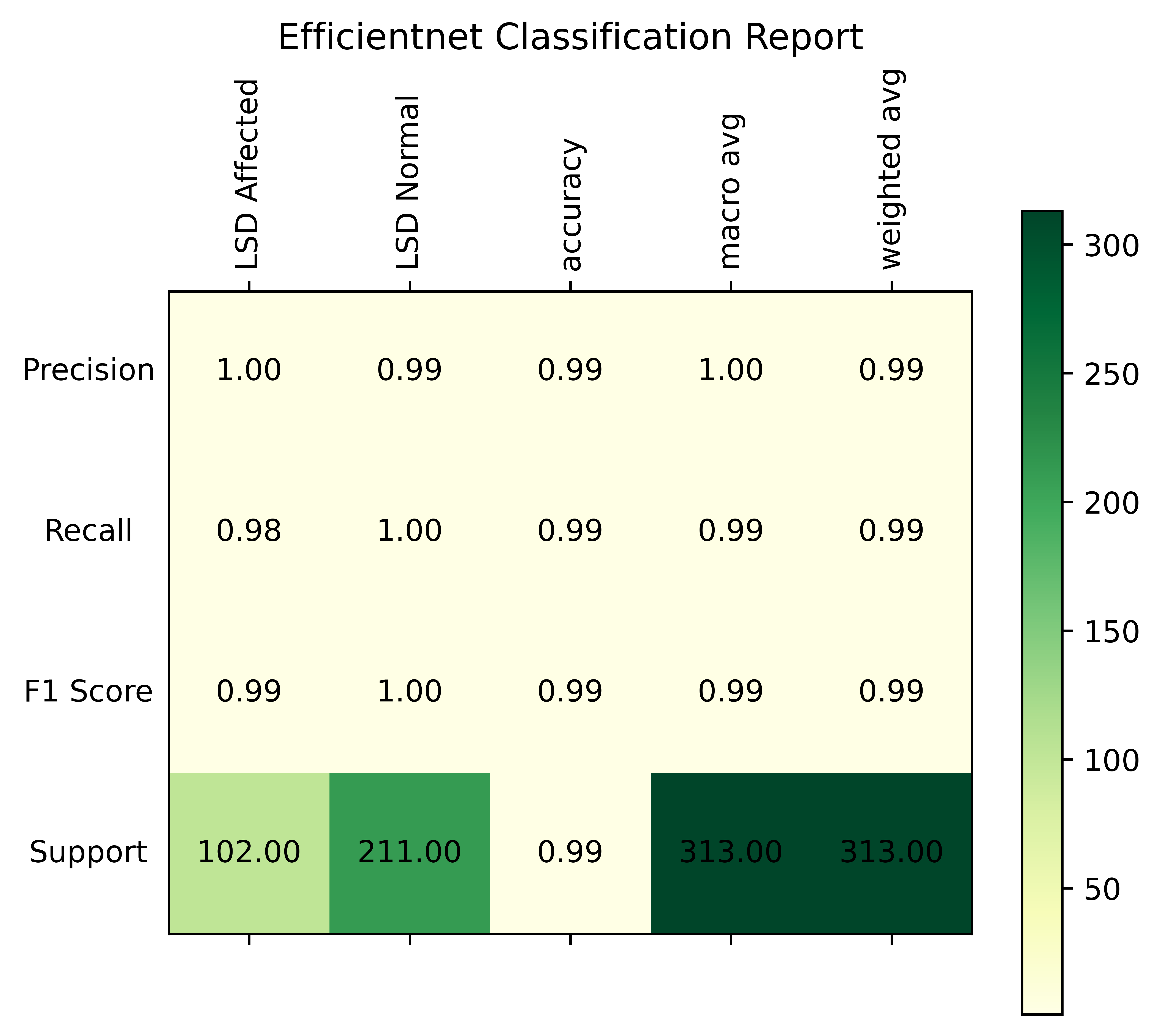}
\caption{LUMPNet Classification Report \label{eff_class_rep}}
\end{figure}

\begin{figure}
\centering
\includegraphics[height = 5.5cm, width = 7cm]{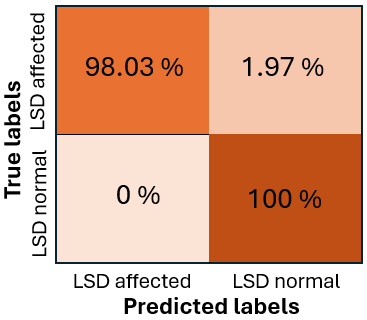}
\caption{LUMPNet Confusion Matrix\label{conf_matrix}}
\end{figure}

\begin{figure}
\centering
\includegraphics[height = 5.5cm, width = 7cm, trim=0cm 0cm 0cm 0.8cm, clip]{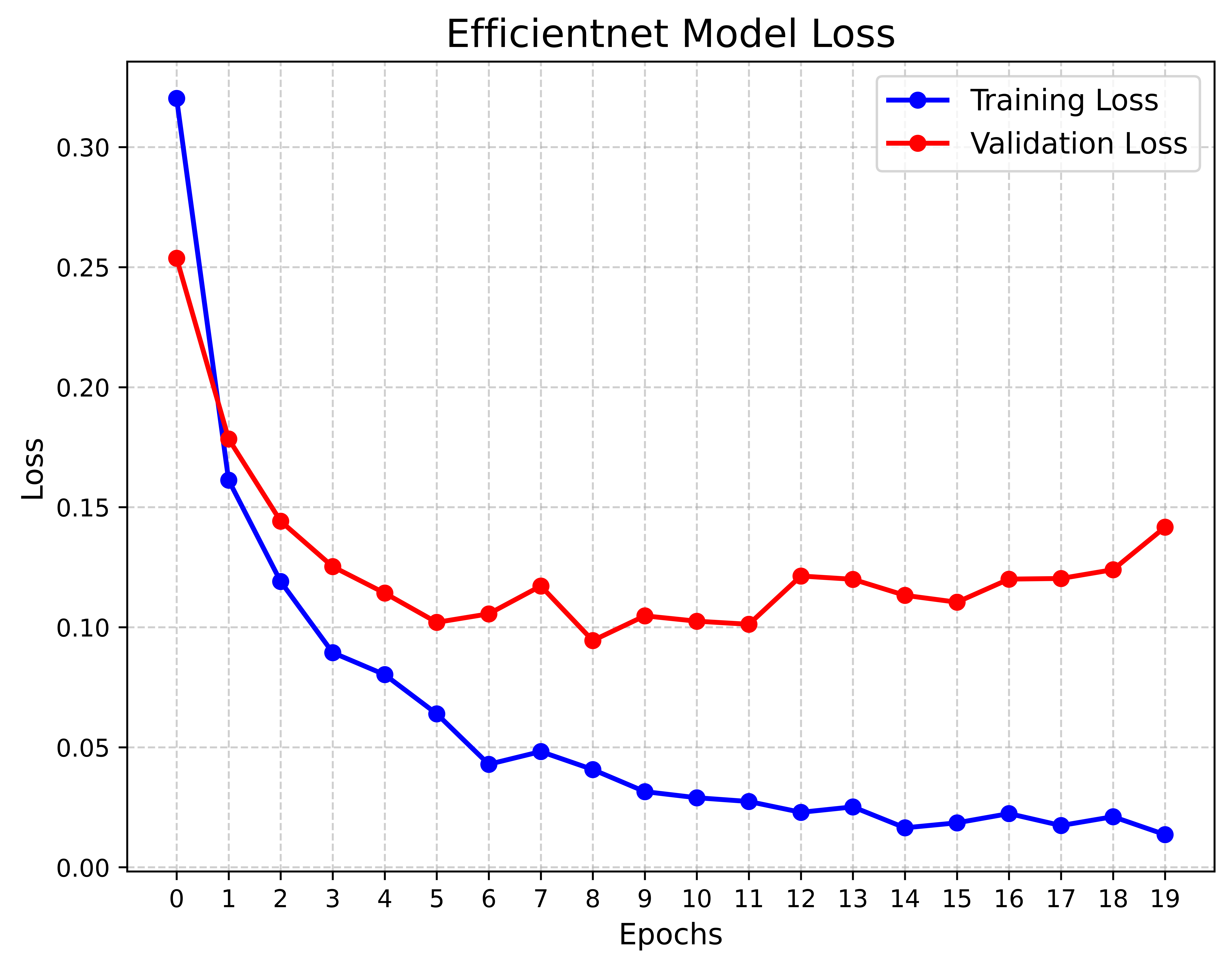}
\caption{LUMPNet Model loss \label{eff_loss}}
\end{figure}

The Precision-Confidence Curve showed a value of 0.978, reflecting the model’s ability to precisely identify cattle with minimal false positives. Meanwhile, the Recall-Confidence Curve scored 0.93, demonstrating that the model’s predictions become increasingly reliable with higher confidence levels. The overall accuracy of the model is 99\%, meaning that nearly all predictions were correct, and both the macro and weighted averages for precision, recall, and F1 score are also 0.99, showing consistently high performance across both classes, even with a slight class imbalance. The LUMPNet model proves to be highly effective in detecting LSD, with exceptional precision and recall, particularly in ensuring that when it predicts a cattle as infected, it is almost always correct, making it a reliable tool for this task. Fig \ref{eff_acc} illustrates the training and validation accuracy of the LUMPNet model over 20 epochs in the task of detecting LSD in cattle. Both training and validation accuracy improve rapidly within the first few epochs, with training accuracy stabilizing around 99\% and validation accuracy fluctuating slightly around 97-98\%. 

\section{Conclusion and Future Work}

The proposed work on detecting LSD in cattle through image data and AI demonstrates a comprehensive and effective workflow that integrates a novel AWDR optimizer with two major ML techniques. This work utilizes YOLOv11 for cattle detection and EfficientNet for accurately identifying LSD symptoms, particularly skin nodules, which are early indicators of the disease. The results show that the LUMPNet model performs exceptionally well, achieving 99\% accuracy, high precision, recall, and F1 scores, confirming its reliability in detecting infected cattle. The systematic approach from dataset preparation to evaluation proves to be transformative for early disease detection, offering significant benefits to livestock management by enabling timely intervention and improving animal health outcomes. This research paves the way for a robust AI-driven solution in veterinary disease management, ensuring more efficient and accurate detection of diseases like LSD in cattle.

LUMPNet model introduces a hybrid deep learning framework that integrates AWDR optimizer and YOLOv11 with the SAM to achieve both object detection and fine-grained segmentation of LSD-affected cattle. Unlike standard YOLOv11, which detects objects using bounding boxes alone, this approach enhances precision by segmenting lesion areas after detection, making it particularly effective for medical image analysis. Additionally, the use of a Dynamic Filtering Layer (DFL) refines the detection outputs by focusing on high-confidence regions, a novel technique that significantly improves accuracy in overlapping or subtle visual cases. The model also leverages enhanced multi-scale feature fusion using custom convolutional layers, C2f blocks, and Spatial Pyramid Pooling (SPPF), which are tailored to detect lesions of case study. On the classification side, EfficientNet is fine-tuned through transfer learning on a domain-specific dataset. The use of compound scaling ensures a balanced trade-off between accuracy and efficiency, allowing for effective lesion classification even with limited training data. A key innovation of the research lies in the seamless integration of detection and classification.  

Future work will focus on expanding the dataset across diverse cattle breeds and environments to improve model generalisability. Additionally, integrating temporal analysis for disease progression tracking and developing a mobile application for real-time, field-based detection will be explored. Further enhancements to the segmentation accuracy using transformer-based models and cross-domain validation with other livestock diseases are also planned to broaden the system’s applicability in veterinary diagnostics.

\section*{Data availability statement}
The datasets generated and/or analysed in the current work are publicly available in the kaggle's repository ``Lumpy Skin Images Dataset'' accessible at \url{https://www.kaggle.com/datasets/warcoder/lumpy-skin-images-dataset}.

\bibliography{references}
\bibliographystyle{IEEEtran}
\end{document}